%% file: main.tex
\documentclass[journal]{IEEEtran}

\usepackage{cite}
\usepackage{amsmath,amssymb,amsfonts}
\usepackage{algorithmic}
\usepackage{graphicx}
\usepackage{textcomp}

\usepackage{url}
\usepackage{pgfplots}
\usepackage{multirow}

\usepackage{array}
\usepackage[caption=false,font=normalsize,labelfont=sf,textfont=sf]{subfig}

\usepackage{stfloats}
\def\BibTeX{{\rm B\kern-.05em{\sc i\kern-.025em b}\kern-.08em
		T\kern-.1667em\lower.7ex\hbox{E}\kern-.125emX}}

\usepackage{balance}

\begin{document}
\title{Intelligent Transportation Systems Using External Infrastructure: A Literature Survey}
\author{\IEEEauthorblockN{Christian Creß$^*$, Zhenshan Bing, Alois C. Knoll}
		\thanks{All authors are with the Technical University of Munich, Department of Informatics, Chair of Robotics, Artificial Intelligence and Real-time Systems, Munich, Germany. \newline
		E-mail: christian.cress@tum.de, bing@in.tum.de, knoll@in.tum.de \newline
		$^*$ Corresponding author.			
		  }}
	
\markboth{Intelligent Transportation Systems Using External Infrastructure: A Literature Survey}%
	{How to Use the IEEEtran \LaTeX \ Templates}
	
\maketitle
	
\begin{abstract}
The main problems in transportation are accidents, increasingly slow traffic flow, and pollution. An intelligent transportation system (ITS) using external infrastructure can overcome these problems. For this reason, the number of such systems is increasing dramatically, and therefore requires an adequate overview. To the best of our knowledge, no current systematic review of existing ITS solutions exists. To fill this knowledge gap, our paper provides an overview of existing ITS that use external infrastructure worldwide. Accordingly, this paper addresses current questions and challenges. For this purpose, we performed a literature review of documents that describe existing ITS solutions from 2009 until today. We categorized the results according to technology levels and analyzed its hardware system setup and value-added contributions. In doing so, we made the ITS solutions comparable and highlighted past development alongside current trends. We analyzed more than 357 papers, including 52 test bed projects. In summary, current ITSs can deliver accurate information about individuals in traffic situations in real-time. However, further research into ITS should focus on more reliable perception of the traffic using modern sensors, plug-and-play mechanisms, and secure real-time distribution of the digital twins in a decentralized manner. By addressing these topics, the development of intelligent transportation systems will be able to take a step towards its comprehensive roll-out.
\end{abstract}
	

\input{introduction}
\input{methodology}
\input{results}

\input{discussion}

\input{conclusion}

\section*{Acknowledgment}
This research was funded by the Federal Ministry for Digital and Transport of Germany in the project Providentia++, FKZ: O1MM19008A. We would like to express our gratefulness for making this paper possible.

\bibliography{literature.bib}
\bibliographystyle{IEEEtran}

\vspace{12pt}

\begin{IEEEbiography}[{\includegraphics[width=1in,height=1.25in,clip,keepaspectratio]{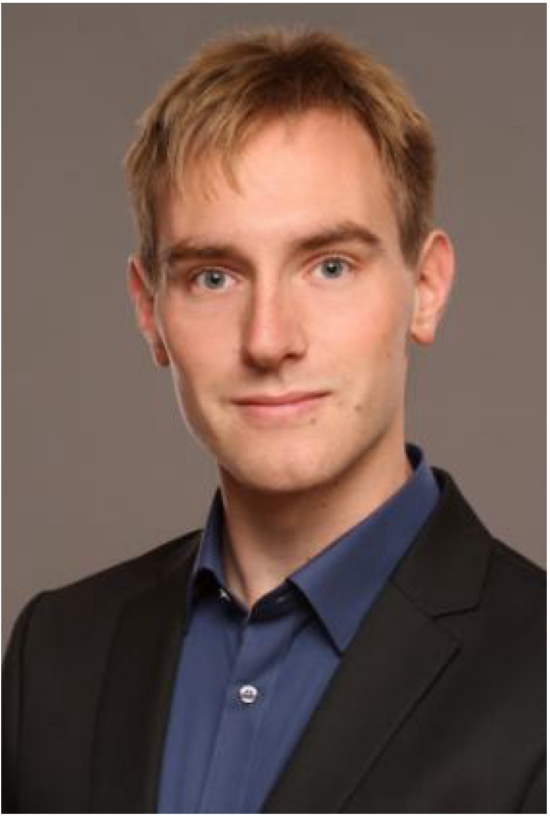}}]{Christian Creß} joined the Chair of Robotics, Artificial Intelligence and Real-time Systems at the Technical University of Munich (TUM), Germany in 2020 as a Research Assistant and Ph.D. student, where he is currently researching computer vision and data fusion for multi-modal sensor systems. He completed his M.Sc. in Applied Computer Science at the University of Applied Sciences Kempten in 2016. The Master's thesis was in the area of computer vision and machine learning. Before starting as Research Assistant at TUM, he worked as a software developer in the industry. His further research interests include computer vision, artificial intelligence, and software architecture. \end{IEEEbiography}

\begin{IEEEbiography}[{\includegraphics[width=1in,height=1.25in,clip,keepaspectratio]{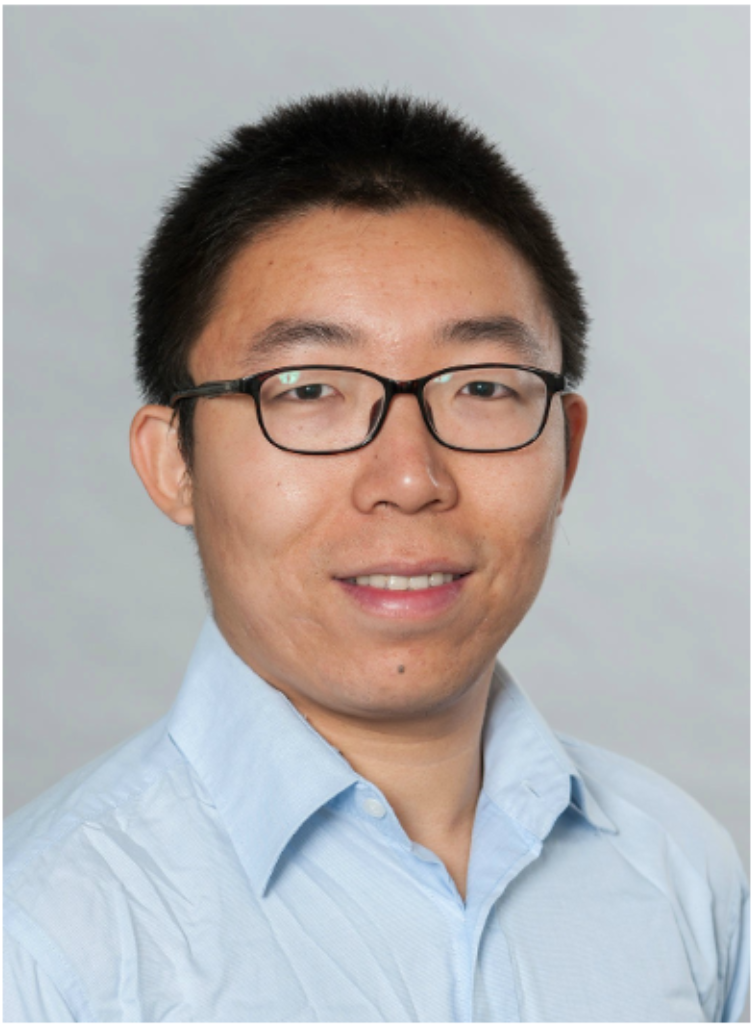}}]{Zhenshan Bing} received his doctorate degree in Computer Science from the Technical University of Munich, Germany, in 2019. He received his B.S degree in Mechanical Design Manufacturing and Automation from Harbin Institute of Technology, China, in 2013, and his M.Eng degree in Mechanical Engineering in 2015, at the same university. Dr. Bing is currently a post-doctoral researcher with Informatics 6, Technical University of Munich, Munich, Germany. His research investigates the snake-like robot which is controlled by artificial neural networks and its related applications.\end{IEEEbiography}

\begin{IEEEbiography}[{\includegraphics[width=1in,height=1.25in,clip,keepaspectratio]{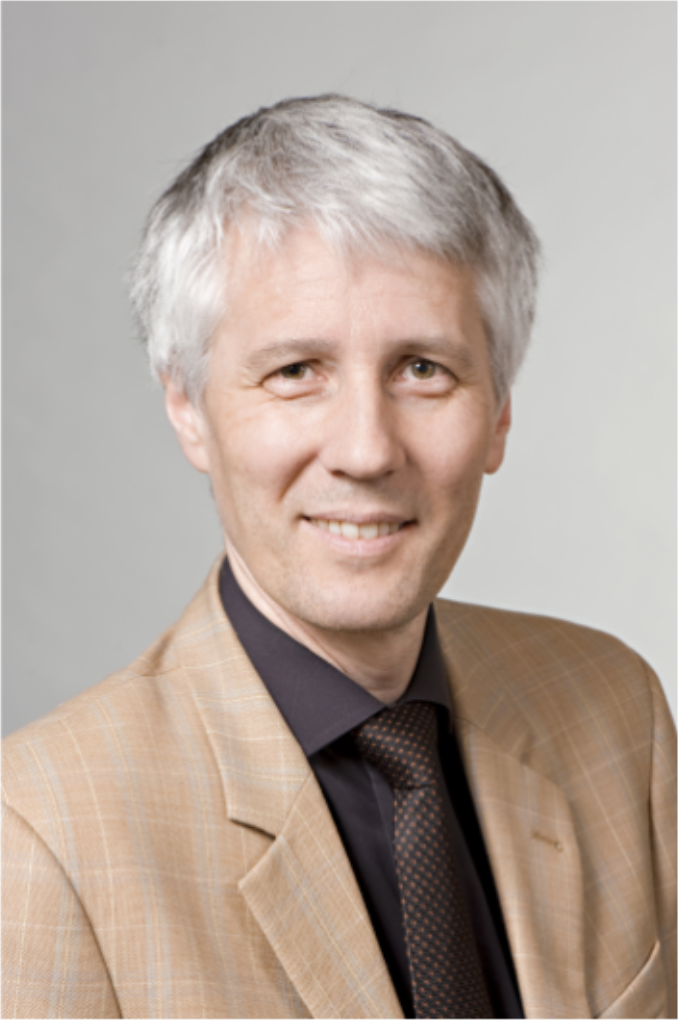}}]{Alois Knoll} (Senior Member) received his diploma (M.Sc.) degree in Electrical/Communications Engineering from the University of Stuttgart, Germany, in 1985 and his Ph.D. (summa cum laude) in Computer Science from Technical University of Berlin, Germany, in 1988. He served on the faculty of the Computer Science department at TU Berlin until 1993. He joined the University of Bielefeld, Germany as a full professor and served as the director of the Technical Informatics research group until 2001. Since 2001, he has been a professor at the Department of Informatics, Technical University of	Munich (TUM), Germany. He was also on the board of directors of the Central Institute of Medical Technology at TUM (IMETUM). From 2004 to 2006, he was Executive Director of the Institute of Computer Science at TUM. Between 2007 and 2009, he was a member of the EU’s highest advisory board on information technology, ISTAG, the Information Society Technology Advisory Group, and a member of its subgroup on Future and Emerging Technologies (FET). In this capacity, he was actively involved in developing the concept of the EU’s FET Flagship projects. His research interests include cognitive, medical and sensor-based robotics, multi-agent systems, data fusion, adaptive systems, multimedia information retrieval, model-driven development of embedded systems with applications to automotive software and electric transportation, as well as simulation systems for robotics and traffic.\end{IEEEbiography}

\end{document}

%% file: introduction.tex
\section{Introduction}
\IEEEPARstart{I}{t} has been reported that more than two million traffic accidents occurred in just the United States from 2005 to 2007. About 94 \% of those accidents were caused by human error \cite{NationalHighwayTrafficSafetyAdministration.2015,Chen.11052018}. Other problems include traffic jams and exhaust pollution \cite{Miles.2006}. Furthermore, 50 \% of the world's population lives in cities, where the problems mentioned are worse. Unfortunately, this situation will become even more critical as the urban population is set to increase by two-thirds by 2050. Therefore, the need to further investigate more advanced traffic optimization solutions is vital \cite{Seuwou.2020}.

\begin{figure}[!t]
	\centerline{\includegraphics[width=\linewidth]{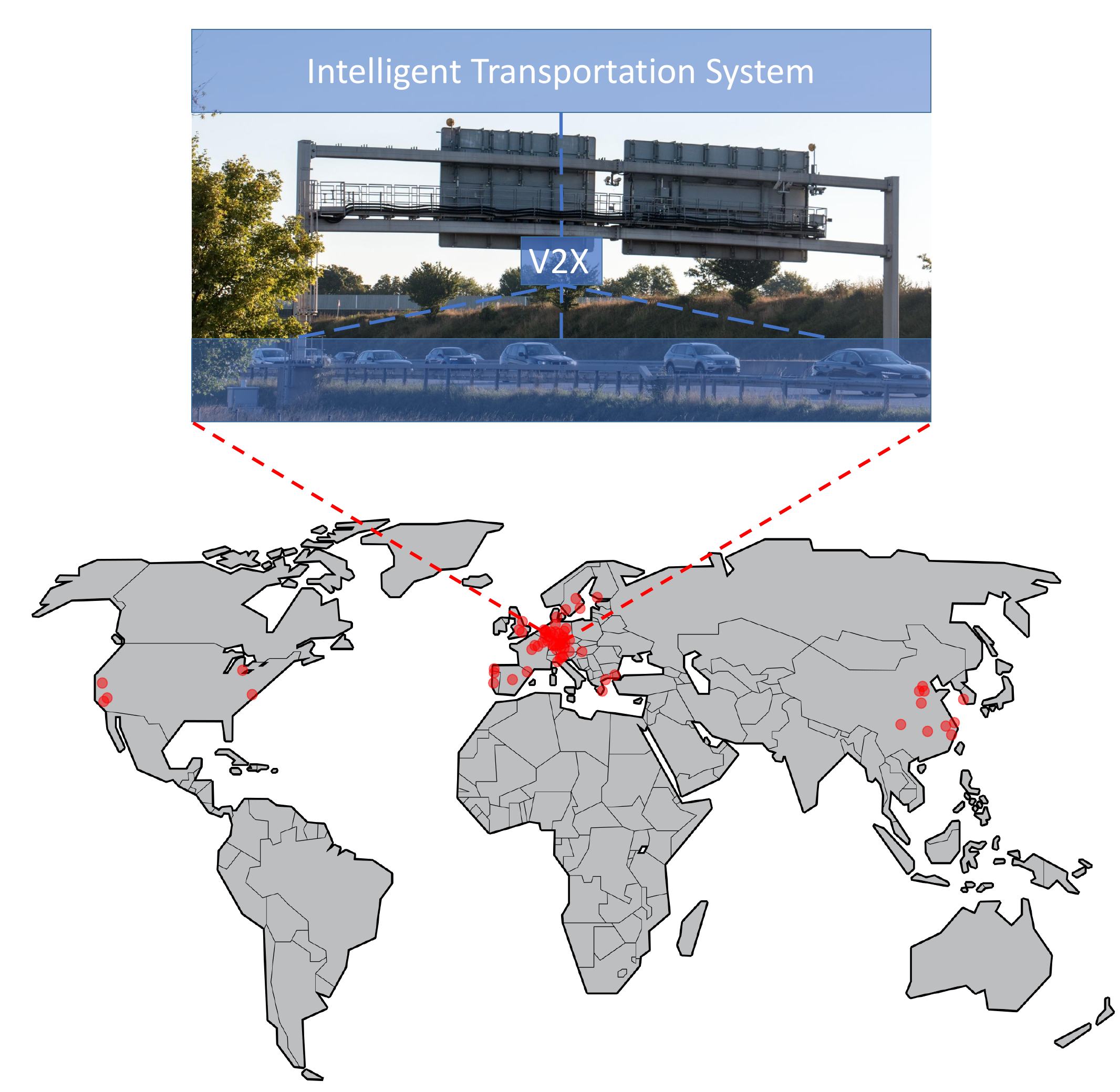}}
	\caption{Intelligent transportation systems are distributed worldwide. In particular, many developments can be seen in North America, Europe, and Asia. The Providentia++ project in Germany shown in the picture is an excellent example of such a system. This survey paper gives a detailed overview of the global ITS solutions using external infrastructure and discovers trends and open research questions.}
	\label{figIntro}
\end{figure}

Intelligent transportation systems (ITS) using external infrastructure are a possible solution for tackling these problems \cite{AmilcareFrancescoSantamaria.2018,Chen.11052018,Lee.2006,Miles.2006,Paier.42015}. As estimated, these systems could prevent approximately 400,000 to 600,000 accidents per year coupled with an in-parallel reduction in travel time of around 42 \% in the United States alone. In addition, modern ITS could reduce the fuel consumption for passenger vehicles by up to 44 \% and for trucks by up to 18 \% \cite{Chen.11052018}. This would result in a reduction of around 15 \% greenhouse gases \cite{Ogie.2017}. Furthermore, ITS could on the one hand implement a real-time synchronization of individual traffic with the public transport system with the support of `Park and Ride'-Parking, and on the other hand exchange data with vehicles that are equipped with intelligent driver assistance systems. This Vehicle-to-Everything technology (V2X) would provide further foresight resulting in an accurate warning of hazards and obstacles.

For the manufacturers of ITS and the automotive industry there is also a business opportunity: in 2014, the worldwide annual turnover was around 30 billion euros for connected driving applications, which increased to 170 - 180 billion euros in 2020. Considering the application `Collaborative Cruise Control' as an example, it enables continuous information exchange between vehicles and the infrastructure and has already been used by Tesla for its feature AutoPilot over a distance in excess of 4 billion miles on the road. Moreover, personalized onboard entertainment, fleet management, and platooning \cite{Sheik.122019} are other influential value-added services of ITS. Thus, these systems could make a valuable contribution to the transportation segment.

To the best of our knowledge, there exists no systematic literature survey of existing intelligent transportation systems using external infrastructure that provides details about hardware and software, and their value-added contributions. Papers \cite{A.Kotsi.2020d,.12112021t,M.Lu.2014,MengLu.2018} and \cite{.12112021c} list ITS in general. However, they do not give technical details nor the exact selection criterion of their listed systems. Therefore, the purpose of this paper is to provide a systematic literature review of the last decade. Furthermore, to improve understanding, we provide a historical summary of such systems. While this review does not cover the specific topics in the field of ITS (e.g. communication, architecture, object detection, or tracking) in depth, it provides an overview of existing ITS solutions, trends in terms of the sensors and software features used, and the current outstanding research questions in the field.

This paper covers documents relating to ITS from 2009 until the present day. In addition, documents published prior to this date have been considered where they relate to the first use of ITS. Then, we categorize the test fields according to their general functionalities. To make the systems comparable, we extract the relevant characteristics in terms of sensors and value-added contributions. Lastly, we highlight and visualize the results and show the trends of the development in the area of ITS. As shown in Figure \ref{figIntro}, ITS are located in North America, Europe, and Asia.

The paper is organized as follows: firstly, section \ref{section:methodology} explains the methods used for collecting and reviewing papers on intelligent transportation systems that use external infrastructure. Section \ref{section:results} shows the results of the literature survey. This section also includes a detailed analysis of each listed ITS. Section \ref{section:discussion} discusses the results, the requirements, and the trends of modern ITS. Lastly, a conclusion and outlook for further research are given in Section \ref{section:conclusion}.

%% file: methodology.tex
\section{Methodology}
\label{section:methodology}
This section describes how we categorized the collected ITS systems according to their features, analyze their hardware and software properties, and discuss their scientific contributions. First, we inspect unsystematically the literature to get an overview of the growing field of ITS. Here, we noticed numerous developments in the English, German and Chinese-speaking countries. Then, to achieve the highest possible coverage of existing literature, we performed a systematic review according to the systematic review method of Kitchenham \cite{B.Kitchenham.2004}: 

\begin{table}[htbp]
	\caption{Development stages of ITS.}
	\begin{center}
		\begin{tabular}{|p{3cm}|p{5cm}|}
			\hline
			\textbf{Category} & \textbf{Definition} \\
			\hline
			Early Days & This category summarizes the concepts, challenges, and successes of the first ITS. \\
			\hline
			Situation-Related Analysis & ITS in this category can analyze and interpret individual traffic situations. Therefore, they can improve safety and traffic flow. \\
			\hline
			Creation of Digital Twins & These systems can create complex digital twins of all traffic participants. They are characterized by high accuracy and real-time capacity. Such systems can have a high impact on autonomous and connected driving.	\\
			\hline		
		\end{tabular}
		\label{tableCategories}
	\end{center}
\end{table}

\begin{enumerate}
	\item Identification of research: We collect documents with the known databases and search engines, namely, IEEE Xplore, Google Scholar, and Google with multilingual keywords. We use `Test Field ITS', `Testbed ITS' and `C-ITS'. Furthermore, we used the German keyword `Testfeld ITS' (Engl.: Test Field ITS) and the Chinese keyword `\includegraphics[height=7pt]{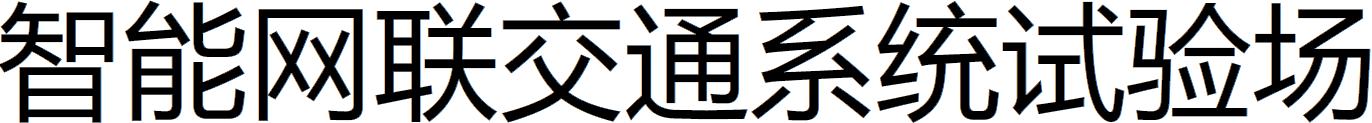}' (Pinyin: Zhìnéng wǎng lián jiāotōng xìtǒng shìyàn chǎng, Engl.: Intelligent Connected Transportation System Test Site). This enables us to achieve higher coverage, especially if it was not published in English. We also consider the related work of the collected publications. With this strategy, we get a pyramid scheme in our research. Unfortunately, a lot of intelligent transportation system projects do not publish their results in peer-reviewed papers. Therefore, we accept in our literature survey both papers and gray literature, e.g. official websites about projects.
	
	\item Study selection: We follow several rules for the assessment or the filtering of collected sources. We select ITS from 2009 to today. For the historical background, we choose exceptionally important first-known test field projects. As an additional criterion, the ITS in this paper has to use external sensors or other infrastructure. The systems must have been implemented in a real-world scenario, systems involving only a conceptual idea were not included in the literature survey.
	
	\item Study quality assessment: After the assessment of all documents, we categorize the ITS solutions according to their stage of development. The authors of \cite{Ogie.2017} proposed the categories `Semi-intelligent infrastructure', `Intelligent or semismart infrastructure' and `Smart infrastructure' in 2017. Here, the categories of the ITS are based on real-time behavior, accuracy, and independent operation without human intervention. However, we suggest in our work the three categories `Early Days', `Situation-Related Analysis' and `Creation of Digital Twins'. These categories should describe the stage of development according to the general features and functionalities of the system. The definition of the categories is described in detail in Table \ref{tableCategories}. 
	
	\item Data extraction: The ITS projects focus on different aspects. For instance, some test fields focus on communication between vehicles and infrastructure, other test fields focus on data fusion or object detection. To make the test fields comparable, we analyze the used sensors, the length of the test stretch, the overall architecture, and the scientific novelty of each ITS. In doing so, we extract the purpose and the technical details of every project from the last decade.  
	
	\item Data synthesis: In the last step of our method, we highlight and visualize the qualitative and quantitative results of the collected ITS. We summarize the derived data in Table \ref{tableITS} and Table \ref{tableValueAddedServices} and indicate the trend of past development in diagrams. Therefore, this synthesis gives a sufficient impression of the current development in the area of ITS as well as anchor points for open research questions.    
\end{enumerate}
    

%% file: results.tex
\section{Results}
\label{section:results}
Based on our proposed method, we analyzed 357 papers about intelligent transportation systems using external infrastructure, which we found using the search engines of IEEE Xplore, Google Scholar, and Google with the mentioned keywords. Since there were more than one million hits for searches performed using Google and Google scholar, only the first 30 search results were taken into account. The number of useful documents for each keyword for each search engine used is listed in Table \ref{tableSearchResults}.

\begin{table}[htbp]
	\caption{Number of relevant documents.}
	\begin{center}
		\begin{tabular}{|c|c|c|}
			\hline
			\textbf{Search Engine} & \textbf{Keyword} & \textbf{Number of results} \\
			\hline
			\multirow{4}{*}{IEEE Xplore$^{\mathrm{*}}$} & Test Field ITS & 78 \\
			 & Test Bed ITS & 14 \\
			 & C-ITS & 180 \\
			 & Testfeld ITS & 0 \\
			 & \includegraphics[height=7pt]{Chinese_Keyword_Intelligent_Networked_Transportation_System_Test_Site.PNG} & 0 \\
			 
			\hline
			\multirow{4}{*}{Google Scholar} & Test Field ITS & 0 \\
			 & Test Bed ITS & 1 \\
			 & C-ITS & 6 \\	
			 & Testfeld ITS & 12 \\
			 & \includegraphics[height=7pt]{Chinese_Keyword_Intelligent_Networked_Transportation_System_Test_Site.PNG} & 3 \\
			 
			\hline
			\multirow{4}{*}{Google} & Test Field ITS & 17 \\
			 & Test Bed ITS & 7 \\
			 & C-ITS & 6 \\	
			 & Testfeld ITS & 25 \\
			 & \includegraphics[height=7pt]{Chinese_Keyword_Intelligent_Networked_Transportation_System_Test_Site.PNG} & 8 \\
			 
			 \hline
			 $\sum$ & & 357 \\
			 \hline
			 \multicolumn{3}{l}{$^{\mathrm{*}}$With topics filter `intelligent transportation systems'.} 			 	 
			\label{tableSearchResults}
		\end{tabular}
	\end{center}
\end{table}

In the next step, we give details about the ITS found which fulfill our mentioned requirements. The feature categories `Early Days', `Situation-Related Analysis' and `Creation of Digital Twins' are described in the respective sections. Table \ref{tableITS} gives a complete overview of the results.  

\subsection{Early Days}
The first evidence of the use of external infrastructure in intelligent transportation systems is found in the 80s and 90s. Here, the projects PROMETHEUS and PATH were decisive in starting development of ITS. In 1988, the PROMETHEUS project was aimed at significantly increasing the safety of future road traffic, as well as improving traffic efficiency and comfort. The subproject Pro-Road concentrated on communication between infrastructure and vehicles with the aim of optimizing road traffic. The aim of relevant sensors (e.g. radar, laser scanner, ultrasonic, and cameras) in the cars of the period was to provide information about the conditions in the environment. The first ideas on hazard warnings (e.g. traffic jams, accidents, ice) arose. It should be noted that the aim of the project was not to dominate drivers' activities with autonomous driving \cite{Williams.1988,Franke.1989}. 

In contrast, the PATH project was aimed at influencing the control of vehicles on highways for optimization of the traffic flow. For this purpose, the highways were equipped with sensors and control units. For instance, in the year 1992, the first vehicle platooning experiments with Doppler radar systems were successfully. However, it was recognized at the time that implementation turn of the millennium would not be possible. The reasons were technical deficiencies and a lack of political acceptance by the population \cite{Shladover.1993,Shladover.2007b}. 

After the first generation of concepts and prototypes, the `next-generation ITS' partially controlled and partially optimized the general traffic flow. Instead of influencing the traffic participants directly as before, they attempted to control vehicles with intelligent traffic light logic \cite{Srinivasan.2006} or with gantries displaying electronically generated messages. The necessary data was collected using induction loops in the road paving. The first ITS of this type came into being in the late 1990s. For instance, in the year 1998, researchers in the DynaMIT (Dynamic Network Assignment for the Management of Information to Travelers) project in Los Angeles created a concept and simulation for an ITS, which was aimed at improving the traffic flow in the event of accidents or other emergencies. In 2006, the concept was implemented using counting loops on the freeway. The sensors sent the data to a computer for further analysis. In summary, the system could measure the traffic load and generate forecasts in real-time. Based on this, the system offered dynamic route guidance, which improved the traffic flow \cite{BenAkiva.1998,Wen.2006}. 

In 2006, the SAFESPOT project researched the interaction between ITS and vehicles with the aim of increasing road safety (e.g. warning about `black spots') and optimizing the traffic flow. ITS were deployed in the following places: 1.) Sweden: Stockholm and Goteborg. 2.) Germany: Dortmund. 3.) Italy: Torino and Brescia-Padova Freeway. 4.) France and Spain: West Europe Test Site. 5.) Netherlands: Corridor Rotterdam-Brabant-Antwerp. Therefore, the locations covered city center, country road and freeway scenarios. In this huge European project, infrastructures were equipped with Road Side Units (RSU), and dynamic cooperative networks were created. These systems enabled communication between vehicles and infrastructure. Based on the shared information, vehicle drivers received achieved enhanced perception of the surrounding vehicles. The ITS created a `local dynamic map' and a `safety margin assistant', which detected critical situations in advance \cite{.23072010,M.Aramrattana.2015}. 

\subsection{Situation-Related Analysis}
The ITS discussed in this section was able to analyze specific traffic situations. Furthermore, the ITS allowed interaction with the individual participants. This circumstance gave definitive a strong value in comparison to the ITS which gave a rough statement about the general traffic flow. In the year 2013, the European cities Bordeaux (France), Copenhagen (Denmark), Helmond (Netherlands), Newcastle (England), Thessaloniki (Greece), Verona (Italy), and Vigo (Spain) were installed with ITS as part of the project Compass4D. A total of 147 Road Side Units and 5 backend offices were set up. The aim was to develop a system to provide warnings for road hazards. The hazards considered were road building sites, slippery roads, wrong-way drivers, and red light violations. In addition, the ITS should optimize fuel consumption in intersection areas to maximize energy efficiency. The ITS introduced a situation-adapted traffic light. As soon as cars arrived the traffic light, they were given a speed recommendation \cite{Vreeswijk.2014}. 

In the year 2012, the project INSIGHT (Intelligent Synthesis and Realtime Response using Massive Streaming of Heterogeneous Data) set up an ITS in Dublin, Ireland. Here, numerous cameras in certain sections, as well as the GPS data of public transport provided information about the current traffic situation in real-time. The information was bundled with messages from Twitter, radio, and possible callers. The purpose of this big data application was to automate the resource management to enable efficient actions in intelligent and networked cities in the event of a disaster \cite{LemmerK..2019,Panagiotou.2016,CORDISForschungsergebnissederEU.08032021}. In 2015, the follow-up project VaVel (Variety, Veracity, VaLue: Handling the Multiplicity of Urban Sensors) extended the INSIGHT system in Dublin and also implemented it in Warsaw, Poland. The ITS collected the same data, as well as data from trams and bicycles. The project had the same goal as the already referred to INSIGHT project \cite{LemmerK..2019,Panagiotou.2016,CORDISForschungsergebnissederEU.08032021b}. 

In the 2016 European project ICSI (Intelligent Cooperative Sensing for Improved Traffic Efficiency), an ITS was implemented in the cities of Lisbon, Portugal, and Pisa, Italy. The research goal was to find a highly scalable distributed ITS for data acquisition and analysis. The objective of its use were traffic flow optimization and reduction of emissions. On an 8 km long section of the A5 freeway in Lisbon, three measuring stations with cameras analyzed the traffic flow, and also detected anomalies (e.g. construction sites, traffic jams, accidents). In parallel in the city center of Pisa, the focus of the project was on a parking guidance system and reducing air pollution. For this purpose, on a stretch of 1 km, 5 measuring stations with sensors were set up. The advantage of these systems is the decentralized horizontal architecture which ensures a high level of scalability \cite{Osaba.2016}. 

\begin{table*}[htbp]
	\caption{Overview of ITS split according to the technology level Early Days, Situation-Related Analysis, and Creation of Digital Twins. The systems without sensors use data from connected vehicles.}
	\begin{center}
		\begin{tabular}{|l|l|l|l|l|l|l|p{0.7cm}|l|l|l|}
			\hline
			\textbf{Year} & \textbf{Name} & \textbf{Place} & \textbf{Length} & \textbf{Camera} & \textbf{Radar} & \textbf{Lidar} & \textbf{Ind. Loops} & \textbf{V2X} & \textbf{Special Inputs} & \textbf{Architecture} \\
			\hline
			1986 & PROMETHEUS & EU & N/A & - & - & - & - & X & - & N/A \\
			1986 & PATH & US & N/A & - & - & - & - & X & - & N/A \\
			1998 & DynaMIT & US & N/A & - & - & - & X & - & - & Centr. \\
			2006 & SAFESPOT & EU & $>$ 20 km & X & - & X & - & X & Weather/RFID/IR & Cloud \\
			\hline
			2013 & Compass4D & EU & $>$ 10 km & X & - & - & - & X & - & Edge-Cloud \\
			2012 & INSIGHT & IE & N/A & X & - & - & - & X & Social Networks & Cloud \\
			2015 & VaVeL & PL & N/A & X & - & - & - & X & Social Networks & Cloud \\
			2016 & ICSI & EU & 9 km & X & - & - & - & - & - & Decentr. \\
			2018 & Fed4Fire+ & BE & 3 km & - & - & - & - & X & - & Edge-Cloud \\
			2017 & DIGINET-PS & DE & 3.7 km & X & - & - & - & X & Weather/road/parking & Edge-Cloud \\
			2016 & ConVeX & DE & 30 km & - & - & - & X & X & - & Cloud \\
			2017 & DRIVE Hessen & DE & 200 km & X & - & - & - & X & - & Edge-Cloud \\
			2012 & Testf. Kassel & DE & 9 km & - & - & - & - & X & - & Edge-Cloud \\
			2018 & Mid. Future Mobility & UK & 300 km & - & - & - & - & X & - & Edge-Cloud \\
			2016 & ECo-AT & AT & N/A & - & - & - & - & X & - & Centr. \\
			2016 & InterCor & EU & $>$ 100 km & - & - & - & - & X & - & Cloud-Edge \\
			2017 & Safe Strip & EU & 79 km & - & - & - & X & X & - & Edge-Cloud \\
			2016 & AUTOCITS & EU & $>$ 17 km & - & - & - & - & X & Internet & Edge-Cloud \\
			2019 & C-Roads CZ & CZ & $>$ 100 km & - & - & - & - & X & - & Edge-Cloud \\
			2014 & SCOOP@F & FR & 2000 km & - & - & - & - & X & - & Edge-Cloud \\
			2017 & SAFARI & DE & 16 km & - & - & - & - & X & - & Cloud-Edge \\
			2018 & Testf. Friedrichshafen & DE & $>$ 5.5 km & - & - & - & - & X & - & N/A \\
			2019 & ZalaZONE & HU & $>$ 100 km & - & - & - & - & X & - & Edge-Cloud \\
			2016 & CARISSMA & DE & N/A & - & - & - & - & X & - & N/A \\
			2014 & AstaZero & SE & 7.4 km & - & - & - & - & X & - & N/A \\
			2015 & Mcity & US & 1 km  & X & - & X & - & X & - & Edge  \\
			2014 & GoMentum Station & US & 30 km & X & - & -  & - & X & - & Edge \\
			2014 & Virginia Smart Roads & US & $>$ 100 km  & - & - & - & X & X & - & Edge \\
			2016 & Chongqing & CN & N/A & - & - & - & - & X & - & N/A \\
			2016 & Shanghai Lingang & CN & $>$ 10 km & X & - & - & - & X & - & Edge \\
			2018 & Chang'an University & CN & 3.5 km & X & X & X & - & X & - & Edge \\
			2018 & Veh. Application North & CN & 3 km & - & - & - & - & X & HD-Map & N/A \\
			2018 & Changsha & CN & 12 km  & - & - & - & - & X & - & N/A \\
			2021 & Sino-German & CN & N/A & - & - & - & - & X & HD-Map & N/A \\
			2018 & Beijing-Hebei & CN & $>$ 10 km & - & - & - & - & X & - & N/A \\
			2017 & Wuxi & CN & $>$ 100 km & - & - & - & - & X & - & N/A \\
			2018 & Zhejiang & CN & 3 km & X & - & - & - & X & - & N/A \\
			\hline
			2017 & MEC-View & DE & $<$ 1 km & X & - & X & - & X & Stereo Cam./Las. Scan. & Edge-Cloud \\
			2019 & Testf. Ger.-Fran.-Lux. & EU & $>$ 100 km & X & - & - & - & X & - & Edge-Cloud \\
			2017 & KoRA9 & DE & 225 m & - & X & - & - & X & - & Edge-Cloud \\
			2018 & 5GMOBIX & EU/Asia & $>$ 100 km & X & X & X & - & X & Weather/road/pedestrian & Edge-Cloud \\
			2017 & Providentia & DE & 3.5 km & X & X & X & - & X & HD-Map/Event-b. Cam. & Edge-Cloud \\
			2014 & AIM Braunschweig & DE & 7 km & X & X & X & - & X & Stereo Cam./Las. Scan. & Edge-Cloud \\
			2016 & Testf. Niedersachsen & DE & 280 km & X & - & - & - & X & Stereo Cam. & Edge-Cloud \\
			2017 & Testf. Düsseldorf & DE & 24 km & X & - & X & - & X & - & Edge-Cloud \\
			2016 & Testf. Bad.-Württem. & DE & $>$ 100 km & X & - & - & - & X & Weather & Edge-Cloud \\
			2009 & Ko-PER & DE & $<$ 1 km & X & - & X & - & X & - & Cloud \\
			2017 & ALP.Lab & AT & 23 km & X & X & X & - & X & - & Edge-Cloud \\
			2020 & SMLL London & UK & 24 km & X & - & - & - & X &  - & Edge-Cloud \\
			2021 & IN2Lab & DE & 2 km & X & X & X & - & X & - & Edge-Cloud \\
			2016 & Testfeld Dresden & DE & 20 km & X & - & - & - & X & - & Edge-Cloud \\
			2019 & HEAT & DE & 1.8 km & - & X & X & - & X & - & Edge-Cloud \\
			\hline
			\multicolumn{11}{l}{$^{\mathrm{*}}$Note: N/A means the information could not be found.} 
		\end{tabular}
		\label{tableITS}
	\end{center}
\end{table*}

Within the scope of the project FED4FIRE+, the ``Smart Highway: V2X Testbed'' was created in the year 2018. The ITS had 7 RSUs over a length of 3 km on the E313 highway in Antwerp, Belgium. The system was also part of the CityLab test bed. The research area of the project was communication between vehicles and other vehicles or infrastructures (V2X). For this, the system used direct antennas and a cloud solution connected via 4G/5G internet. The system achieved latency times of 10-15 ms. Instead of using a variety of external sensors, this infrastructure only depended on the data of the vehicle. For data-intensive applications, the ITS used high-performance computers. The FED4FIRE+ project presented a new approach for the collocation of `multi-access edge computing' platforms with the support of RSUs. This has improved the Quality of Service in infotainment services for vehicles on the highway \cite{FED4FIRE+.2018,SlamnikKrijestorac.192021}. 

The current Diginet-PS test field in Berlin, Germany, covers the scenarios of federal roads and secondary roads in city areas. The ITS uses cameras, weather sensors, and other traffic sensors (e.g. for measurement of the traffic flow, parking spaces, pollution) for detection of the environment. On the RSU, local data processing is implemented as edge computing. Furthermore, a cloud system receives the results from the RSU. The main purposes of the cloud system are to store, analyze and predict test bed wide data. Based on this ITS, the researchers focus on the development of services and applications to increase traffic safety for automated driving. For instance, the applications evaluate parking space utilization, analyze the road conditions, and determine traffic situations \cite{LemmerK..2019,BMVI.08032021,DAILaborTUBerlin.08032021}. 

In 2016, the ConVeX project created two ITS solutions. The first ITS was located on the A9 Freeway near Nuremberg, Germany. Over a length of 30 km, it was equipped with 6 RSUs. The second ITS was located in Rosenheim, Germany, and had two RSUs. The data inputs of the systems were induction loops from existing infrastructure and the sensor information of the connected cars. Data processing took place in the RSUs and in a traffic control center. The research focus of the project was value-added services, as well as the V2V and V2X communication. The project implemented the value-added services ``Emergency Electronic Brake Light'', ``Do Not Pass Warning'', ``Blind Spot Warning'' and ``Vulnerable Road User Collision Warning''. To influence the traffic flow, the system sent the relevant information directly to the road user or to the electronic message gantries. Furthermore, the ITS notified vulnerable groups (e.g. cyclists, pedestrians) with a smartphone app \cite{ConvexProjekt.2018}. 

The ITS DRIVE-Testfeld Hessen (Dynamic Road Infrastructure Vehicle Environment) is located close to Frankfurt, Germany. The overall goal is the intelligent management of traffic and construction sites. Furthermore, the ITS focuses on new techniques for traffic data acquisition with the use of V2V and V2X communication as well as data fusion between vehicles and infrastructure. The DRIVE-Testfeld Hessen implemented intelligent applications which should support automated driving as well as the general traffic flow. Therefore, the test bed has more than 120 RSUs. The applications have been traffic hazard warnings, shockwave damping, and Green Light Optimal Speed Advisory (GLOSAR) \cite{LemmerK..2019,BMVI.08032021,CROADSGermany.14032021,Schneider.2019}. The mentioned ITS was a platform for several projects. The project simTD set up many RSUs and an ITS Central Station. The value-added services of this project were obstacle warning, electronic brake lights, and warnings of approaching emergency vehicles and construction sites \cite{Commmoveddg1.b.3.2015}. Then, the project CVIS (Cooperative Vehicle-Infrastructure Systems) focused on V2V and V2X communication issues for enhanced driver awareness in an inter-urban environment \cite{.05022020,M.Lu.2014,M.Aramrattana.2015}. Last, but not least, the project DRIVE C2X performed a comprehensive assessment of cooperative systems based on field tests \cite{M.Lu.2014,.18112021}. 

Another ITS exists in the city center of Kassel, Germany. Here, the traffic lights are equipped with 15 RSUs, and an existing traffic management center is used. The system allows priority treatment of public transport and emergency vehicles at traffic lights. Furthermore, the ITS warns road users about approaching emergency vehicles. For this purpose, the system uses the so-called DENM (Decentralised Environmental Notification Messages) for V2V and V2X communication. As value-added services, the researchers have developed a forecast for the green phase and alternative route control \cite{LemmerK..2019,BMVI.08032021,kassel.de:DeroffizielleInternetauftrittderStadtKassel.14032021,Emmett.12112021}. 

The Midlands Future Mobility test field in the United Kingdom is used to research connected and autonomous vehicles. Over a length of 300 kilometers, the ITS is located between the cities Coventry, Birmingham, and Solihull. Therefore, the ITS covers city center, country road, and motorway scenarios. The test field use RSU, which offers various value-added services. For instance, the services could be in-vehicle signage, warnings of roadworks as well as GLOSAR. Current research results on the ITS have been achieved in the area of cooperative perception for 3D object detection using infrastructure sensors \cite{MidlandsFutureMobility.08032021,Arnold.2020}. 

In the context of the European Cooperative ITS Corridor of 2013, vehicles communicated with the support of the infrastructure on the Rotterdam-Frankfurt-Vienna route. The Eco-AT project (2016) in Austria, a follow-up project of Telematik, set up a central ITS station as well as a number of RSUs. The architectural approach was more centralized. The researchers generated the necessary specifications. The applications were decentralized environmental notification message applications (e.g. road works warnings), aggregation of cooperative awareness messages, in-vehicle information (e.g. speed restrictions), and intersection safety (e.g. red light detection) \cite{Paier.42015, ASFINAG.14032021}. 

In 2016, the overall goals of the European project InterCor (Interoperable Corridors) were improving the safety, the traffic flow as well as increasing the cybersecurity of the ITS system itself. The project aimed to link several ITS corridors to exchange innovation and best practices in solving common problems. Hence, the project included ITS operations in Belgium, France, the Netherlands, and the United Kingdom. The environments of the ITS in Belgium, the Netherlands, and France were freeways. Furthermore, the ITS in France was an extension of the SCOOP@F project, which is introduced below. The environment in the United Kingdom was urban roads and motorways. The project involved the installation of RSUs which allowed V2X communication. For instance, the project achieved the value-added services of road works warnings, in-vehicle signage, onboard signaling of hazardous and unexpected events, multi-modal cargo optimization, green light optimization, probe vehicle data, parking information services for trucks, and tunnel management \cite{InterCor.19112021,A.Didouh.2020,G.Crockford.2018,M.Msahli.2015}. 

The project SAFE STRIP (SAFE and green Sensor Technologies for self-explaining and forgiving Road Interactive aPplications) implements an unconventional new approach for increasing traffic safety. The ITS consists of intelligent infrastructure sensors in the form of strips with networking capabilities. They are installed the road pavement surface and analyze the individual situations. For instance, the system measures environmental parameters (e.g. temperature, ice, ambient light), passing vehicles (speed, lateral position, vehicle type) as well as pedestrian crossings. Furthermore, these sensors communicate bidirectionally with vehicles as well as external infrastructure (e.g. traffic management centers). For test and implementation, the project has installed an ITS in a 9 km section near Rovereto on the A22 freeway in Italy as well as in a 70 km section in the Attiki Odos urban freeway in Athen, Greece. Here, the project has placed sensors on 39 toll barriers in Athens and several strip sensors as well as 5 RSUs in Rovereto \cite{M.Gkemou.2019,F.Biral.2019,SafeStrip.25112021}. 

The deployment of ITS solutions in Europe was aimed at in the European Project AUTOCITS in the year 2016. The overall goal of the project is to boost the role of ITS as a catalyst for the implementation of autonomous driving. Therefore, several ITS in Lisbon (Portugal), Madrid (Spain), and Paris (France) were created. The test bed in Madrid consisted of 15 RSUs which were set up on a busy freeway over a length of 10 km. Over a total length of 8 km on the freeway and urban road, the ITS in Lisbon used 5 RSUs. Last, but not least, on an urban highway near Paris, the ITS consisted of 1 RSU. The project has achieved various day 1 services. For instance, road works warning, slow or stationary vehicle warning, as well as the providing of information about weather conditions has been implemented \cite{J.E.Naranjo.2018,J.E.Naranjo.2020,C.Premebida.2018,.23112021b}. 

In the year 2019, the project C-Roads CZ focused on the implementation of ITS units as well as the testing and evaluation of their functionalities. The features of the ITS were situated-related analysis (e.g. road works warning, in-vehicle information, slow and stationary vehicles, traffic jam ahead warning, hazardous location notification). For testing under real conditions, seven test fields have been built in the Czech Republic. All these ITS solutions have been equipped with RSUs which communicate with vehicles and a central processing system \cite{Z.Lokaj.2020}. 

In the project SCOOP@F, an ITS was deployed on a nationwide scale in 2014. For this, the project equipped 2000 km of road with RSUs as well as 3000 vehicles with onboard units for V2X communication. The RSUs were connected to traffic management centers. The main data sources of the test field were the vehicles themselves. They delivered data about their position, speed, and direction. Furthermore, vehicles received information about unexpected events (e.g. construction, slippery road, reduced visibility) from external sources. The main goals of the ITS were to increase safety and to improve travel quality. The researchers implemented a decentralized communication system to alert road users about unexpected events. In addition, the enhancement of the coverage of VANETs, a secure cloud environment for connected vehicles as well as a novel detailed security protocol were achieved \cite{Aniss.2016,J.P.Monteuuis.2017,H.Labiod.2015,F.Haidar.2015b,G.Wilhelm.2020,H.Fouchal.2018}. 

Another important test bed is SAFARI in Berlin. The ITS covers the scenarios main road, secondary road, and a junction of a freeway. The overall goal of the project is the exchange of information and the dynamic update mechanism of the dynamic maps, which are basic requirements for automated driving. In detail, the project researched the handling of construction sites and unexpected events for automated and connected driving. The test field uses cloud platforms as well as mobile edge computing for the V2X processing \cite{LemmerK..2019,BMVI.08032021,BMVI.19062018}. 

In the test bed Friedrichshafen, traffic lights are equipped with RSUs and V2X. Overall, the ITS covers the traffic scenarios of federal roads, country roads, city centers, and pedestrian zones. The research topic is the information exchange between vehicles and infrastructure, as well as between vehicles themselves. The current research project on the test field is Alfried \cite{LemmerK..2019,BMVI.08032021,.02122021}. 

The project ZalaZONE in Hungary has a cordoned-off proving ground as well as an ITS up to the border with Austria. The ITS is applied to a highway scenario. In the cordoned-off proving ground, it covers the scenarios of smart cities, high and low-speed handling as well as high-speed freeways and rural roads. The objectives are testing and research in the field of autonomous and connected vehicles. For this, the test field is equipped with V2X communication technology \cite{Szalay.2019}. 

Another cordoned-off test facility for re-enacting and simulating traffic scenarios is implemented in the CARISSMA project in Ingolstadt. The test facility offers 4000 square meters for researching V2X communication, accident detection, accident consequences reduction as well as the automated driving functions \cite{TechnischeHochschuleIngolstadt.20142021}. 

The next cordoned-off ITS is AstaZero (Active Safety Test Area). It was set up in 2014 in Sweden, and it is an open environment where vehicle OEMs, research institutes, and universities perform development and research. The research topics are new active safety functions for road vehicles. The test field contains a city area, 700 m multilane road, 5.7 km long rural road, and 1 km high-speed area \cite{Jacobson.2015}. 

Further cordoned-off ITS are Mcity and GoMentum Station, located in Michigan and California, United States. The proving grounds include highway scenarios, rural roads with tunnels, and urban settings, for example, roundabouts, intersections, and railroad crossings. The sensors used, e.g. cameras or lidars, are connected to Road Side Units via 5G, V2X technology, or fiber optics. These setups allow testing of autonomous and connected driving \cite{LiuTianyangi.2017, Mcity.18032022, GoMentumStation.24052021}. 

The system Virginia Smart Roads is a cordoned-off test bed that provides advanced-vehicle testing with Roadside Units (DSRC). The highlights are V2X communication, use of paving sensors, and differential GPS broadcast. Furthermore, the test bed provides 75 weather-making towers, which can generate snow, fog, and rain. The closed test bed covers highway as well as rural road sections. For testing under real conditions, the Virginia Smart Roads are extended by the Virginia Automated and Connected Corridors with 70 miles of interstate highways. There are 56 RSUs for V2X communication \cite{LiuTianyangi.2017, VirginiaTech.VirginiaSmartRoads}. 

In addition to the test beds in Europe and the U.S., there are also numerous cordoned-off test beds in China. There is a cordoned-off test bed for connected and autonomous driving up to Level 4 in Chongqing. The test bed covers expressways and urban areas. Here, the test bed generates weather scenarios like rain and fog. The research focus is V2X communication with 5G for autonomous driving and Advanced Driver Assistance Systems \cite{hatchip.com.,autotesting.net.2021,diandong.com.2018,YiMao.2019}. 

The cordoned-off test bed in Shanghai Lingang covers more than 10 km of highways and urban roads. It provides special scenarios, such as rainfall and tunnels. The main points are V2X communication and perception for autonomous driving. The test bed uses 1 GPS differential base station, LTE-V, DSRC Units, intelligent traffic lights, and cameras \cite{hatchip.com.,autotesting.net.2021,iot.ofweek.com.2019,diandong.com.2018}. 

The cordoned-off test bed at Chang'an University covers a length of more than 3.5 km. The setup enables perception testing with a camera, radar and lidar. The V2X communication is realized with LTE, LTE-V, Wi-Fi, and EUHT. The purpose of the test bed is the development of autonomous driving functions, e.g. emergency collision warning, blind-spot warning, or identification of traffic lights \cite{LiXiaochi.2019,hatchip.com.}. 

The National Intelligent Connected Vehicle Application North is another cordoned-off test bed in China. On a length of 3 km, the proving ground covers urban expressways, intersections, tunnels, and a rain-fog generator. Furthermore, the facility contains HD maps as well as communication with 5G. Another cordoned-off test field is in Changsha. The test bed has a length of 12 km. The communication in the scenarios' highway, urban area, and rural road are realized with 5G network \cite{autotesting.net.2021}. 

The cordoned-off test bed Sino-German aims to be 200 square kilometers of a huge smart city for testing intelligent connected driving. The facilities contain an HD map and 5G-based V2X technology. With this equipment, the test bed enables more than 116 test scenarios in the categories of safety, efficiency, information service, new energy applications, communication, and positioning capability tests. For example, the actual tests are blind-spot warning, emergency vehicle approaching warning and prioritization, pedestrian collision warning, and traffic light optimization (GLOSAR) \cite{diandong.com.2018}.  

Some test beds in China are set up on public roads, in addition to the cordoned-off area. For example, the Beijing-Hebei Demonstration Zone contains the cordoned-off test beds in Shunyi and Haidian as well as a public road test bed in Yizhuang. The purpose is to test autonomous driving functions. The test bed in Haidian has a length of 4.8 km, and in Yizhuang of 8 km. The test beds offer V2X communication via LTE and 5G network. The scenarios covered are urban roads and expressways \cite{autotesting.net.2021,waytous.com.2021}. 

Huge activities relating to intelligent transportation systems are located in Wuxi. First, a cordoned-off test bed was opened in 2017: over a stretch of 3.53 km, the test bed includes urban roads, rural roads, and highways. The focus is on intelligent traffic management with connected vehicles \cite{diandong.com.2018}. Second, the public roads in Wuxi are equipped with V2X technology. For this, general vehicles and public transportation systems are connected with intelligent traffic lights and electronic message gantries via LTE. The overall goal is increased efficiency in the transportation segment \cite{Hua.08072018,autotesting.net.2021}. 

Another facility on public roads is in Zhejiang. There, the test bed provides high-definition cameras and V2X technology. The real-time communication between vehicles and a command center is realized with 34 infrastructure-based LTE-V or 5G units. Its purpose is also to increase efficiency \cite{diandong.com.2018,autotesting.net.2021}. 

\subsection{Creation of Digital Twins}
Autonomous vehicles are being developed by Google \cite{EricoGuizzo.13022021}, Tesla \cite{Dikmen.op.2017} as well as many other automotive manufacturers. They can tackle some of the known problems in the transportation segment. According to the current approaches in autonomous driving, the primary responsibility, which is associated with replacing human perception and control, lies on vehicle manufacturers. However, the ITS on the infrastructure can create high-precision digital twins. The vehicles can use this data as ground truth information. Therefore, the responsibility can be re-balanced between vehicles and infrastructure. Thus, high precision ITS can be an essential enabler for autonomous driving \cite{Gopalswamy.06022018}. This section presents several ITS solutions which are aimed at achieving the referred to properties above.

In the year 2017, the MEC-View project implemented an ITS in the city center of Ulm, Germany. Across roads and an intersection, the system contained 8 cameras, 4 stereo cameras, 4 laser scanners as well as 8 lidars. For object prediction and data fusion, the sensors were connected to a server. Vehicles communicate with the server via 5G. In addition, a 3D-HD map of the relevant road sections was created. The project aimed to create a digital twin of road traffic in real-time \cite{.22032021,.2021o}. 

The digital test bed ``Germany-France-Luxembourg'' was created in 2016. The focus was on the testing of automated and connected driving functions. The highlight of this project was the challenges of national border crossing. As part of the mentioned test bed, the project ITeM (ITS Testfeld Merzig) in Merzig, Germany, covered the environments of city and rural roads. The research on this ITS was on connected driving functions via C2V and C2X. For traffic perception, the infrastructure was extended with an RSU, C2X technology, and cameras. The RSU communicated with vehicles and a centralized server. The project has implemented applications for hybrid access technologies and safety functions \cite{LemmerK..2019,BMVI.08032021,Informationsplattformfur5GIndustriellesInternet.14112019,.12112021b}. 

In 2017 on freeway A9 in Germany, researchers of the project KoRA9 (Kooperative Radarsensoren für das digitale Testfeld A9. English: Cooperative radar sensors for the digital test field A9) installed 10 novel 77 GHz radar sensors on five masts along the freeway at separations of 45 m. Furthermore, 2 Road Side Units were installed in a gantry bridge for the V2X communication. The project goal was to evaluate the potential of seamless sensory recording of traffic with radars. For this purpose, the infrastructure sensors performed automotive radar applications. The obtained data streams were merged into a digital twin. The researchers have analyzed the digital twin and have made it available to third parties via the developed cloud platform. Furthermore, they have developed services for infrastructure support during overtaking maneuvers and traffic jam warnings \cite{BMVI.08032021,BMVI.08032021b}. 

\begin{figure*}[!t]
	\centerline{\includegraphics[width=\linewidth]{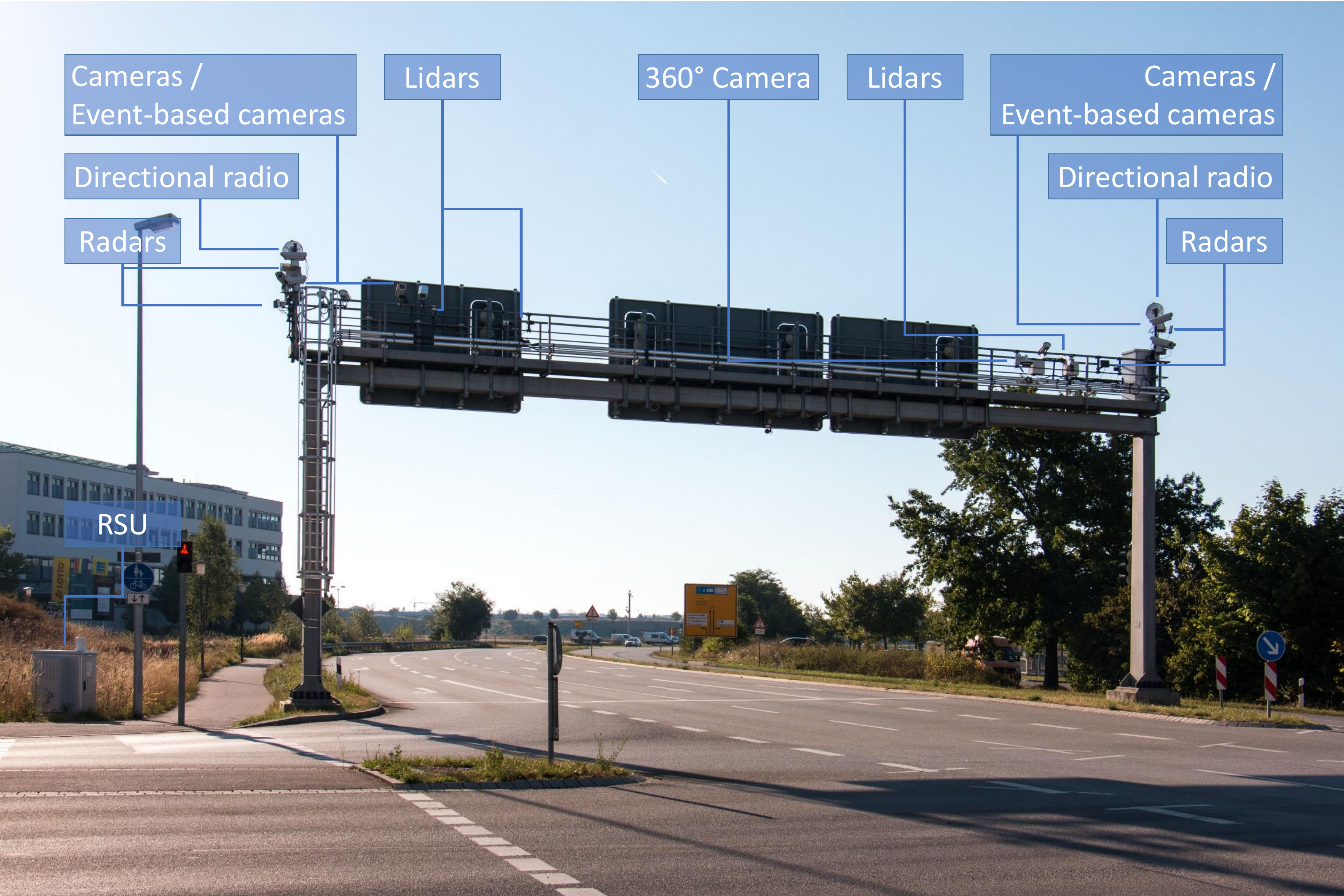}}
	\caption{The ITS Providentia++ has 7 measuring stations with 75 sensors over a length of 3.5 km. The overall goal of the system is to create a digital twin of each traffic participant in real-time. The intersection shown is a measuring station in an urban area. It contains an RSU and includes the sensor types camera, 360° camera, event-based camera, radar, and lidar. Data exchange with the other measuring points is realized via directional radio.}
	\label{figProvidentia}
\end{figure*}

With the European Project 5GMOBIX (5G for cooperative \& connected automated MOBIlity on X-border corridors), several ITS in Europe and Asia were created or reused in 2018. The project showcased the benefits of 5G technology for cooperative, connected, and automated mobility. The Spain-Portugal cross-border ITS between Vigo and Porto focused on automated driving maneuvers. It was equipped with 2 MEC nodes, 2 center clouds, 5 RSUs, 4 traffic radars, and 2 pedestrian detectors. The ITS Greece-Turkey researched platooning and the use of external sensors. The external infrastructure contains 6 edge computers, 2 clouds, 4 RSUs and 2 lidars. The ITS in Eindhoven-Helmond in the Netherlands focused on cooperative collision avoidance with extensive use of cameras. Therefore, the ITS used 2 MEC servers, 2 cloud systems, and more than 50 cameras. The ITS in France were located in Versailles and Paris. They focused on infrastructure-assisted driving with 4 MEC nodes, 1 cloud, 2 lidars as well as 3 cameras. The ITS in Berlin and Stuttgart, Germany, researched RSU-assisted platooning and surround-view generation. Therefore, the system used 9 RSUs, 1 cloud system, 18 cameras, and 10 environmental condition sensors. The ESPOO urban ITS in Finland focused on extended sensor processing as well as remote driving. The ITS was set up with 2 MEC nodes, 1 cloud system, 1 lidar, and 1 camera. The researchers of the Jinan ITS in China focused on cloud-assisted advanced driving, cloud-assisted platooning as well as remote driving. The equipment used was 5 RSU nodes, 1 cloud, 5 lidars, 6 cameras, and 5 radars. Last, but not least, the cordoned-off ITS Yeonggwang in South Korea focuses on remote driving as well. The system used 1 cloud system \cite{A.Serrador.2021,.23112021}. 

With the support of external infrastructure, the Providentia project created a digital twin of each traffic participant in real-time in the year 2017. This digital twin should provide distant view for autonomous vehicles as well as vehicles with conventional driver assistance systems. For this purpose, an ITS was built on the A9 freeway in Munich, Germany. The ITS contained 2 measurement stations on gantry bridges which collected data on the traffic. Each station was equipped with a RSU, 4 cameras, and 4 radars. Furthermore, a backend computer realized the global data fusion as well as the connection to the 5G network and the internet. The sensors could detect the individual traffic participants and sent the data to the RSU. Then, the RSU tracked the detections over several time steps. The tracks were merged into a local digital twin for each RSU. In the end, the backend receives the local twins of the RSU and has combined them again into a global digital twin across all measuring points. With this, the ITS of Providentia has created digital twins which combine the advantages of radar and camera from different perspectives \cite{Hinz.,Krammer.17062019,V.Lakshminarasimhan.2018}. In 2020, the follow-up project Providentia++ extended the ITS into urban areas. This enabled the analysis of complex traffic scenarios in intersections and pedestrian crossings. In total, the test bed had a length of 3.5 km and contained 7 measuring stations with 75 sensors (e.g. cameras, event-based cameras, radars, and lidars). Figure \ref{figProvidentia} shows the hardware setup of Providentia++ on a public road intersection in an urban area. At the time of writing this paper, the researchers of Providentia++ have established a stable data stream of the high-precision digital twin for the general public.\footnote{Livestream: \url{https://innovation-mobility.com/projekt-providentia/}} To the best of our knowledge, this feature is unique in the current development of ITS \cite{PROVIDENTIAA9TestfeldfurautonomesFahrenunddigitaleErkennungvonFahrzeugen.26112021}. 

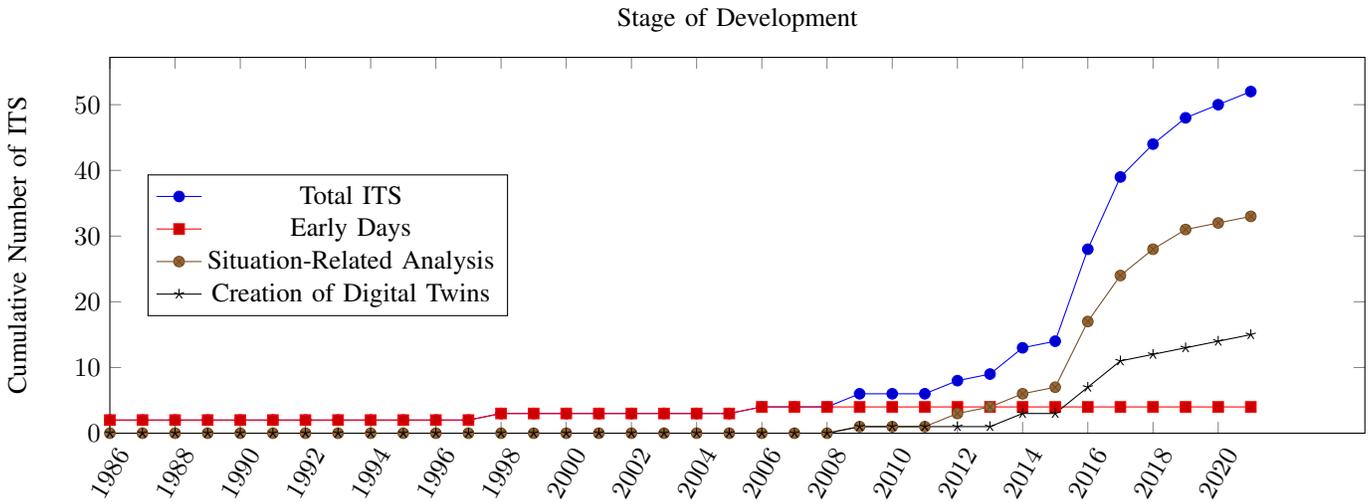
\begin{figure*}[!t]
	\centerline{\input{diagram_feature_trend_cumulative}}
	\caption{In this paper, we have categorized intelligent transportation systems according to their stage of development. It is noticeable that there has been a large increase in ITS systems since 2015. Nevertheless, the development of systems for Situation-Related Analysis seems to have slowed down significantly since 2019. In addition, there is still a lot of potential for development of digital twin systems.}
	\label{figFeatureTrendITS}
\end{figure*}

In 2014, the ITS AIM (Application Platform for Intelligent Mobility) started in Braunschweig, Germany. The ITS infrastructure is designed as a long-term platform for research projects in the area of autonomous and connected driving in an urban environment. The test bed has a length of 7 km and is equipped with RSUs for V2X communication across 35 intersections with traffic lights. The ITS enables information sharing between traffic lights, vehicles as well as other infrastructure. For perception, the several RSUs are connected with radar, camera, stereo-camera, and lidars or laser scanners. The goal of the overall system is the delivery of digital twins with corresponding video scenes under real-time conditions \cite{LemmerK..2019,KnakeLanghorst.2016,.29112021,T.Frankiewicz.2013c, Frankiewicz.2011}. 

The project Testfeld Niedersachsen (Englisch: Lower Saxony test bed) in Germany is an extension of the mentioned AIM test bed in Braunschweig. The test bed runs over a total length of 280 km of freeways and connects the cities Hannover, Hildesheim, Braunschweig, and Wolfsburg. On a 7.5 km long section, the ITS has 71 measuring stations for high-precision recording of traffic situations. The test bed is equipped with stereo cameras and 3 RSUs which provides V2X communication as well as a backend server. The ITS aims to offer an open research platform that delivers ground truth data for autonomous and connected driving. The overall goals are reduction of accidents, improvement of perception, and optimization of the traffic flow. The researchers have realized the value-added services' maintenance vehicle warning, in-vehicle signage as well as lane management \cite{DLRVerkehr.05032021,Koster.2018,.2020,CROADSGermany.12112021}. 

The digital test bed ``Düsseldorf'' in Germany runs on a length of 20 km on the A57 freeway from the Meerbusch junction near Krefeld via the Kaarst junction onto the A52 into downtown Düsseldorf. Therefore, the ITS covers the scenarios of freeway, tunnel, bridge, city center, and parking. The test bed is expanding with several projects. The project KoMoD researched the interaction between infrastructure and vehicles in the scope of connected and automated driving functions. The ITS used vehicles as mobile sensors for the detection of obstacles on the road (e.g. pedestrians, wrong-way drivers). The project has developed the value-added services of autonomous parking, prioritization of public transport systems and cooperative traffic light \cite{www.komodtestfeld.org.06082020}. Then, the follow-up project KoMoDnext should prepare numerous sections of the ITS for autonomous driving (level 4). The project has designed and implemented a high-resolution environment detection on the A57. This component has merged stationary and vehicle-generated data \cite{komodnext.19082020,BMVI.08032021}. Last, but not least, the project ACCorD expanded the corridor for New Mobility Aachen-Düsseldorf by an additional 4 km. The central component of the project activity was the generation of a digital twin. The highlight was the extensive usage of the lidar technology: The ITS contained 68 measurement stations. Each station has 2 cameras and 2 lidars so 136 lidars were set up. This data has was processed in various RSU and a central database, which offered the collected data to other research activities and simulation \cite{ACCorD.10112021b}. 

The ITS Testfeld Autonomes Fahren Baden-Württemberg (English: Test Bed Autonomous Driving Baden-Württemberg), which is located in Karlsruhe and Heilbronn, uses cameras, several RSUs and a backend server. The test bed allows a sensor extension with radars and lidars. Based on camera object tracking, the ITS aims to generate digital twins of the traffic. With this low-cost sensor setup, real-time recording in various traffic situations (e.g. traffic with a vehicle, bicycle, and pedestrian) is implemented. The purpose of the infrastructure is the testing of vehicle systems for automated and connected driving under real traffic conditions \cite{Fleck.2018,BMVI.08032021,.12112021j}. For instance, current research on the test field is the CARAMEL project. The project researched the challenges in the cybersecurity of modern vehicles. For this, CARAMEL applies advanced artificial intelligence and machine learning techniques \cite{.22112021,J.Casademont.2020}. 

In the year 2009, the project Ko-PER of the research initiative Ko-FAS aimed to increase road safety. The goal was a precise creation of a digital twin of the traffic participants in real-time. Therefore, intersections in Germany were equipped with cameras and lidars. The intersection in Ulm contained 2 laser scanners and 2 cameras, in Alzenau 8 laser scanners and 7 cameras, and last, but not least, the intersection in Aschaffenburg contained 14 laser scanners and 10 cameras. The data streams of the sensors on each intersection were combined locally. Then, the fusion result was combined with vehicle information. Therefore, V2X communication technology was implemented too. In doing so, the researchers created a digital twin of the traffic \cite{Rauch.062011,Rauch.062012,ForschungsinitiativeKoFASVerbundprojektKoPER.}. 

As part of the project ALP.Lab - Austrian-Public Test Track Highway A2, an ITS near Graz was implemented in 2017. The test bed covered a distance of 23 km and was equipped with more than 100 sensors. For the creation of digital twins, the ITS used cameras, radars, and lidars. The goals of the ITS were an extension of the vehicles' perception range and to offer ground truth data for plausibility checks of vehicle sensor data. With this, the ITS offers comprehensive testing functions for automated driving and driver assistance systems. Interestingly, the researchers have achieved tracking of vehicles for more than 2 km \cite{ALP.LabGmbH.16032021,Seebacher.112019}. 

The Smart Mobility Living Lab in London is a real-world environment for testing and developing future mobility solutions. For this purpose, the 24 km long test bed has 276 cameras, 40 RSUs, and 2 data centers, which are connected via fiber optics. The RSU provides a V2X communication service. With this hardware, the ITS collects high resolution and high confidence information about the road traffic. The overall goal is supporting the development of autonomous vehicles up to a market-ready stage \cite{SmartMobilityLivingLab:London.22032021}. 

As an extension of the already referred to cordoned-off test facility CARISSMA, the project IN2Lab offers a real-world test bed in Ingolstadt, Germany. The ITS has a total length of approximately 2 km. There is a measurement station with a 4.5 m high mast every 200 m. 13 measuring stations are connected to a central computing unit via optical fiber. Each measuring station has an RSU, cameras, lidars, radars as well as V2X communication modules. The purpose of the test field is researching and testing autonomous vehicles \cite{Agrawal.972021,BranchenNewsfurTopEntscheiderderAutomobilindustrie.12112021}. 

In the scope of the project Testfeld Dresden (English: Dresden test bed), a cooperative ITS is deployed in the urban and suburban areas of Dresden, Germany. The test field provides various driving scenarios (e.g. urban roads, connection to freeways, complex junctions, tram tracks, bus lanes, and separate cyclists lanes) over a 20 km road section. Therefore, the test bed contains several camera systems, 10 RSUs for research and 25 RSUs for day-to-day measurements. Furthermore, the ITS has a backend that hosts a centralized cloud service. The overall goal is the testing of automated and connected driving in a realistic traffic environment. The project has proposed a reference platform for heterogeneous ITS communication as well as several value-added services (e.g. GLOSAR, probe vehicle data, and a cooperative lane change) \cite{S.Strobl.2019c,R.Jacob.2018,Elektromobilitat.01112021}. 

In addition, further ITS solutions for connected and autonomous driving were implemented in numerous cities in Germany. For instance, the project HEAT (Hamburg Electric Autonomous Transportation) installed 2019 a test bed with a length of 2 km in Hamburg. The ITS contained several RSUs, radars, and lidars. The purpose of the sensors was to support the environment perception of an autonomous electric minibus in a real-world scenario. Therefore, the research goal of the HEAT project was the networking of an automated fleet of minibusses with traffic infrastructure. The researchers on this project successfully achieved this goal \cite{LemmerK..2019,HochbahnHamburg.08032021,.2021b}.

%% file: diagram_feature_trend_cumulative.tex
\begin{tikzpicture}
	\begin{axis}[
		title = {Stage of Development},
		ylabel= {Cumulative Number of ITS},
		xmin = 1986, ymin = 0,
		xtick = {1986, 1988, ..., 2021},
		ytick = {0, 10, ..., 60},
		scaled ticks=false,
		ticklabel style={
			/pgf/number format/fixed,
			/pgf/number format/1000 sep={}
		},
		x tick label style={rotate=60},
		legend style={at={(0.03,0.5)},anchor=west},
		scale only axis,
		width=0.92\textwidth,
		height=5cm,
		]
		
		\addplot coordinates {
			(1986, 2)
			(1987, 2)
			(1988, 2)
			(1989, 2)
			(1990, 2)
			(1991, 2)
			(1992, 2)
			(1993, 2)
			(1994, 2)
			(1995, 2)
			(1996, 2)
			(1997, 2)
			(1998, 3)
			(1999, 3)
			(2000, 3)
			(2001, 3)
			(2002, 3)
			(2003, 3)
			(2004, 3)
			(2005, 3)
			(2006, 4)
			(2007, 4)
			(2008, 4)
			(2009, 6)
			(2010, 6)
			(2011, 6)
			(2012, 8)
			(2013, 9)
			(2014, 13)
			(2015, 14)
			(2016, 28)
			(2017, 39)
			(2018, 44)
			(2019, 48)
			(2020, 50)
			(2021, 52)
		};
		\addlegendentry{Total ITS}
		
		\addplot coordinates {
			(1986, 2)
			(1987, 2)
			(1988, 2)
			(1989, 2)
			(1990, 2)
			(1991, 2)
			(1992, 2)
			(1993, 2)
			(1994, 2)
			(1995, 2)
			(1996, 2)
			(1997, 2)
			(1998, 3)
			(1999, 3)
			(2000, 3)
			(2001, 3)
			(2002, 3)
			(2003, 3)
			(2004, 3)
			(2005, 3)
			(2006, 4)
			(2007, 4)
			(2008, 4)
			(2009, 4)
			(2010, 4)
			(2011, 4)
			(2012, 4)
			(2013, 4)
			(2014, 4)
			(2015, 4)
			(2016, 4)
			(2017, 4)
			(2018, 4)
			(2019, 4)
			(2020, 4)
			(2021, 4)
		};
		\addlegendentry{Early Days}
		
		\addplot coordinates {
			(1986, 0)
			(1987, 0)
			(1988, 0)
			(1989, 0)
			(1990, 0)
			(1991, 0)
			(1992, 0)
			(1993, 0)
			(1994, 0)
			(1995, 0)
			(1996, 0)
			(1997, 0)
			(1998, 0)
			(1999, 0)
			(2000, 0)
			(2001, 0)
			(2002, 0)
			(2003, 0)
			(2004, 0)
			(2005, 0)
			(2006, 0)
			(2007, 0)
			(2008, 0)
			(2009, 1)
			(2010, 1)
			(2011, 1)
			(2012, 3)
			(2013, 4)
			(2014, 6)
			(2015, 7)
			(2016, 17)
			(2017, 24)
			(2018, 28)
			(2019, 31)
			(2020, 32)
			(2021, 33)
		};
		\addlegendentry{Situation-Related Analysis}
		
		\addplot coordinates {
			(1986, 0)
			(1987, 0)
			(1988, 0)
			(1989, 0)
			(1990, 0)
			(1991, 0)
			(1992, 0)
			(1993, 0)
			(1994, 0)
			(1995, 0)
			(1996, 0)
			(1997, 0)
			(1998, 0)
			(1999, 0)
			(2000, 0)
			(2001, 0)
			(2002, 0)
			(2003, 0)
			(2004, 0)
			(2005, 0)
			(2006, 0)
			(2007, 0)
			(2008, 0)
			(2009, 1)
			(2010, 1)
			(2011, 1)
			(2012, 1)
			(2013, 1)
			(2014, 3)
			(2015, 3)
			(2016, 7)
			(2017, 11)
			(2018, 12)
			(2019, 13)
			(2020, 14)
			(2021, 15)
		};
		\addlegendentry{Creation of Digital Twins}

	\end{axis}
\end{tikzpicture}

%% file: discussion.tex
\section{Discussion}
\label{section:discussion}
Intelligent transportation systems using external infrastructure offer great potential and the possibility to advance connected and autonomous driving. To provide these future uses, the sensors, and services of ITS are becoming more demanding and complex. To achieve political acceptance, the ITS must also take privacy and security aspects into account. Based on the presented literature survey, the following research questions are identified: 1.) What is the general purpose trend of ITS? 2.) What is the necessary sensor setup? 3.) Which value-added services already exist, and where is there still a need for research? 4.) Which non-functional requirements for ITS have to be considered? These questions have not yet been adequately answered, and so we would like to discuss them in this section.

\subsection{General Purpose Trend}
The complexity of the tasks of ITS has increased significantly. In the Early Days, ITS systems improved the safety and efficiency of the traffic flow. Therefore, the systems allowed a general statement to be made about the current traffic flow. Furthermore, prototypes of the first V2X were developed. This development started in 1986 and ended in the noughties. In the next step, the ITS of the category Situation-Related Analysis analyzed and interpreted individual traffic situations. Furthermore, these ITS solutions could share their results with other traffic participants. Consequently, these systems contributed to more safety and better traffic flow. This category grew exponentially from 2010 to 2018 and then stagnated. Presumably, the functional scope of these systems is now no longer sufficient. Since 2012, more and more test beds have been developed which can generate highly accurate digital twins of road traffic. They allow a complete, high granularity analysis of the traffic. So far, there is no sign of stagnation. As future ITS solutions should give control recommendations to autonomous vehicles, this current development is necessary. Figure \ref{figFeatureTrendITS} shows the cumulative number of test beds in each category. The trend of ITS is more and more in the creation of high-precision digital twins. The development of test beds in the category of Situation-Related Analysis appears to be stagnating.

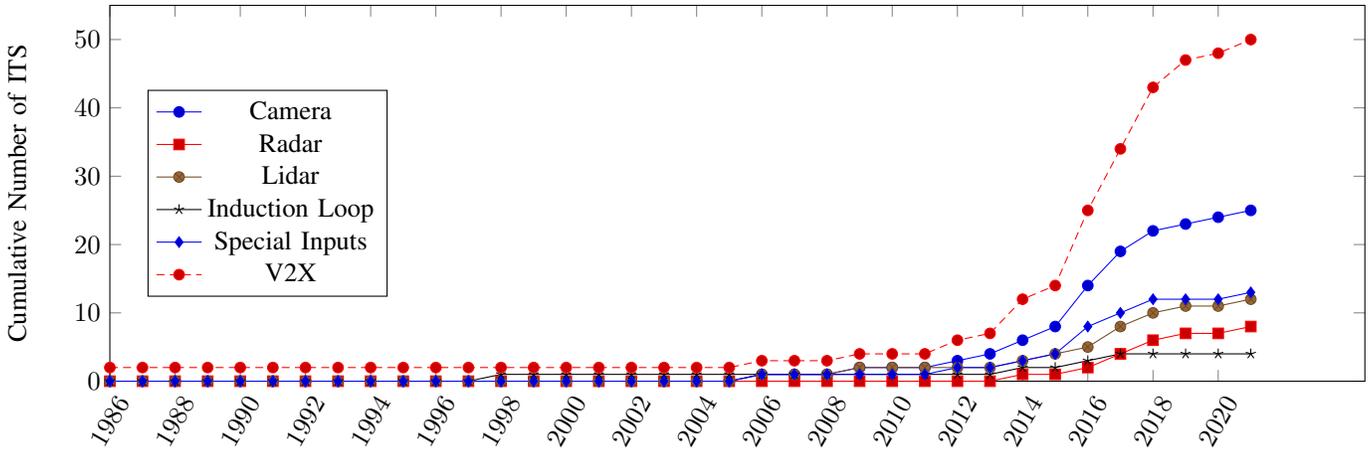
\begin{figure*}[!t]
	\centerline{\input{diagram_sensor_trend_cumulative}}
	\caption{To ensure the optimal perception of the traffic, modern systems rely on various types of sensors in addition to V2X technology. In particular, cameras and lidars are becoming increasingly important. Interestingly, classic induction loops are hardly considered for current systems.}
	\label{figSensorTrendITS}
\end{figure*}

\subsection{Sensor Setup}
The first step in the process chain of ITS is the sensors. They collect information about the real world. Thus, they have the main impact on the precision and quality of the services. In the first test beds, the sensors for analyzing the traffic flow were induction loops. Unfortunately, induction loops on their own do not allow an individual situation analysis. Therefore, cameras have been used in most ITS since 2006. Cameras can create a bird's-eye view of what is happening on the road so that they allow complex situation analysis. The weak points are obvious: the camera does not deliver reliable data at night or in extreme weather conditions (e.g. strong rain, fog, or snow). For compensation, the modern ITS for creating digital twins more and more frequently uses a combination of cameras with radars, lidars, or other special inputs (e.g. IR cameras, stereo cameras, event-based cameras, weather sensors, or external sources from the internet). These hardware setups can perceive the environment in any weather conditions. Interestingly, since the beginning of ITS development, V2X technology has been constantly present. Figure \ref{figSensorTrendITS} includes the full trend of sensor setups in ITS. 

Modern vehicles contain cameras, radars, and lidars too. Strictly speaking, the moving vehicles could be part of the ITS sensor setup. However, the ITS uses extensive V2X technology, and the sensors of vehicles receive too little attention in the overall fusion for generating digital twins. The integration of such  `mobile sensor stations' into modern ITS needs further research. In particular, the early sensor fusion with vehicles should be more considered.

\subsection{Value-Added Services}
The capture of traffic data and the creation of digital twins must not become an end itself. Otherwise, ITS would be useless to society. Instead, ITS has to offer meaningful services which support people in their daily life. For instance, an ITS can decrease maintenance costs, reduce pollution, improve transfer speed and protect human life \cite{Ogie.2017,Seuwou.2020}. Additionally, to cover the high costs of installation and maintenance, ITS services also have a potential for high sales \cite{Datta.,Sheik.122019}. Table \ref{tableValueAddedServices} shows the value-added services in the mentioned test beds.

\begin{table*}[htbp]
	\caption{Overview of the value-added services of ITS.}
	\begin{center}
		\begin{tabular}{|l|c|c|c|c|c|}
			\hline
			\textbf{Value-added service} & \parbox{2.5cm}{\textbf{Warning in operational maneuvers}} & \parbox{2.5cm}{\textbf{Warning in \\road conditions}} & \parbox{2.5cm}{\textbf{Warning in \\traffic situations}} & \parbox{2.5cm}{\textbf{Controlling of \\vehicles}} & \parbox{2.5cm}{\textbf{Controlling of \\traffic flow}} \\
			  & \parbox{2.5cm}{Black spot, emergency electronic brake light, do not pass, road collision} & \parbox{2.5cm}{Construction sites, slow or stationary vehicle, slippery road} & \parbox{2.5cm}{Wrong-way driver, traffic jams, accidents, red light violation, emergency vehicles arriving} & \parbox{2.5cm}{Platooning, remote driving, cooperative lane change, optimize overtaking maneuvers, autonomous parking} & \parbox{2.5cm}{ \vspace{0.1cm} Parking guidance, alternative route control, speed advisory on traffic lights, in-vehicle signage, shockwave damping, priority for public transport or emergency vehicles, lane management \vspace{0.1cm} } \\
			\hline
			PROMETHEUS & - & X & X & - & - \\
			PATH & - & - & - & X & - \\
			DynaMIT & - & - & - & - & X \\
			SAFESPOT & X & - & X & - & - \\
			\hline
			Compass4D & - & X & X & - & X \\
			INSIGHT & - & - & - & - & - \\
			VaVeL & - & - & - & - & - \\
			ICSI & - & X & X & - & X \\
			Fed4Fire+ & - & - & - & - & - \\
			DIGINET-PS & - & X & X & - & X \\
			ConVeX & X & - & - & - & - \\
			DRIVE Hessen & X & X & X & - & X \\
			Testfeld Kassel & - & - & - & - & X \\
			Midlands Future Mobility & - & X & - & - & X \\
			ECo-AT & - & X & X & - & X \\
			InterCor & - & X & X & - & X \\
			Safe Strip & - & - & - & - & - \\
			AUTOCITS & - & X & - & - & - \\
			C-Roads CZ & - & X & X & - & X \\
			SCOOP@F & - & X & X & - & - \\
			SAFARI & - & X & X & - & - \\
			Testfeld Friedrichshafen  & - & - & - & - & - \\
			ZalaZONE & - & - & - & - & - \\
			CARISSMA & - & - & - & - & - \\
			AstaZero & - & - & - & - & - \\
			MCity & X & - & - & X & - \\
			GoMentum Station & X & - & - & X & - \\
			Virginia Smart Roads & X & N/A & N/A & X & N/A \\
			Chongqing & X & - & - & X & - \\
			Shanghai Lingang & X & - & - & X & - \\
			Chang'an University & X & - & - & X & - \\
			Vehicle Application North & N/A & - & - & N/A & - \\
			Changsha & N/A & - & - & N/A & - \\
			Sino-German & X & - & X & X & X \\
			Beijing-Hebei & X & - & - & X & - \\
			Wuxi & X  & X & X & - & X \\
			Zhejiang & X & X & X & - & X \\
			\hline
			MEC-View & - & - & - & - & - \\
			Testfeld Ger.-Fran.-Lux. & - & - & - & - & - \\
			KoRA9 & - & - & X & X & - \\
			5GMOBIX & - & - & - & X & - \\
			Providentia & - & - & - & - & - \\
			AIM Braunschweig & - & - & - & - & - \\
			Testfeld Niedersachsen & - & X & - & - & X \\
			Testfeld Düsseldorf & - & - & X & X & X \\
			Testfeld Bad.-Württem. & - & - & - & - & - \\
			Ko-PER & - & - & - & - & - \\
			ALP.Lab & - & - & - & - & - \\
			SMLL London & - & - & - & - & - \\
			IN2Lab & - & - & - & - & - \\
			Testfeld Dresden & - & - & - & X & X \\
			HEAT & - & - & - & X & - \\
			\hline
			
			\multicolumn{6}{l}{$^{\mathrm{*}}$Note: N/A means the information could not be found.} 
		\end{tabular}
		\label{tableValueAddedServices}
	\end{center}
\end{table*}

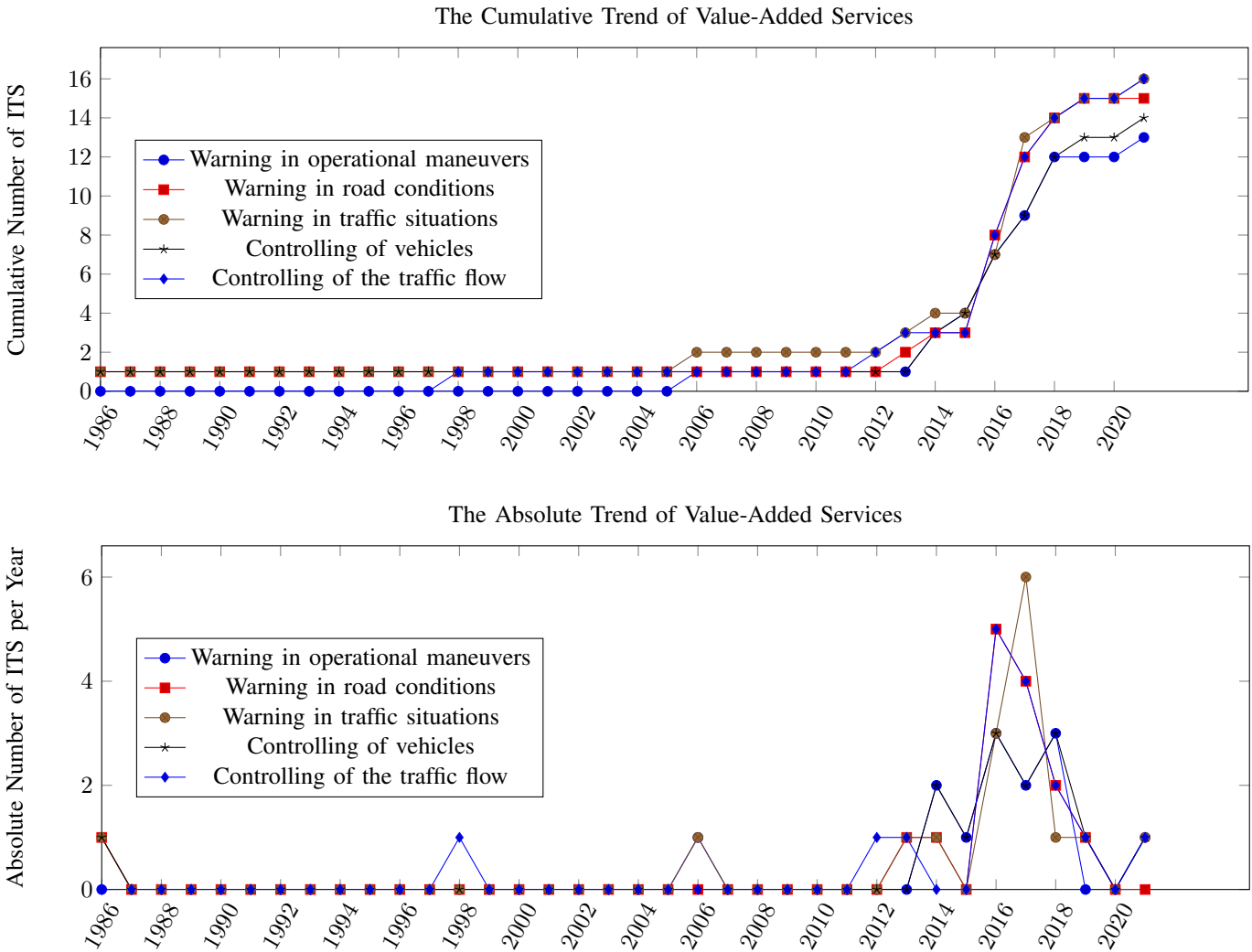
\begin{figure*}[!t]
	\centerline{\input{diagram_value_added_service_trend_cumulative}}
	\vspace{0.5cm}
	\centerline{\input{diagram_value_added_service_trend_absolute}}
	\caption{Intelligent transportation systems can generate meaningful Value-Added Services for the general public. From 2015 in particular there has been a sharp increase in systems that have warned of bad road conditions and dangerous traffic situations. However, this trend declined significantly from 2018 onwards. The trend for systems that offer support in controlling vehicles has grown steadily since 2012. The growth continues today. However, a similarly strong development trend with systems for warning of road conditions or other hazards cannot be identified.}
	\label{figValueAddedServices}
\end{figure*}

Unfortunately, Table \ref{tableValueAddedServices} and the Fig. \ref{figValueAddedServices} highlight a deficiency in the development of operational maneuvers and controlling vehicle services. We noticed that this area is mainly covered by the cordoned-off testbeds in China and the United States. Unfortunately, these areas are not covered in the modern systems on public roads. For this reason, further research should focus on the development of these services on public roads. However, ITS projects \cite{BMVI.08032021b,komodnext.19082020,.12112021j,ALP.LabGmbH.16032021,SmartMobilityLivingLab:London.22032021,Agrawal.972021} create digital twins to support the development of autonomous driving, there are still gaps in using this information in market-ready products. This step is a necessity to cover the high costs of such systems, as well as to provide real added value to the users of ITS solutions.

\subsection{Non-functional Requirements}  
To help the decision-making process of specific maneuver actions of traffic participants, the ITS must be extremely reliable throughout its operation, even in extreme scenarios such as natural disasters. Therefore, ITS are critical infrastructure and needs several mechanisms for safety and cybersecurity. To fulfill the safety requirement, the systems must-have self-diagnostic and self-healing  functionalities. In such circumstances, the ITS would still be safe to operate if malfunctions occurred \cite{Miles.2006,Ogie.2017,Lee.2016}. 

In addition, mechanisms are needed to protect the ITS against cyberattacks. In particular, with potentially fatal consequences, the RSU is very exposed to this kind of attack. Research gaps still exist. For instance, communication between vehicles and test beds is largely unencrypted. This makes the system very vulnerable. To tackle these problems, the researchers have developed communication encryption with cloud support \cite{Chen.11052018}, priority-based authentication techniques \cite{Sheik.122019} and real-time secure communication without overloading the network with a security overhead \cite{Joshi.2017}. Fortunately, some test bed projects have addressed cybersecurity too: the project InterCor has introduced a cyber-physical blockchain cryptographic architecture \cite{A.Didouh.2020} and the project SCOOP@F a secure cloud environment as well as a novel security protocol \cite{J.P.Monteuuis.2017,G.Wilhelm.2020}. Furthermore, the project CARAMEL using the Baden-Württemberg ITS includes activities relating to advanced artificial intelligence and machine learning techniques to fulfill cybersecurity requirements \cite{.22112021}. However, it seems the cybersecurity challenges have not yet been definitively solved. In particular, the secure transmission of high volumes of data, as occurs with detailed digital twins, still seems to require research. Therefore, further research on these security aspects is needed.  

In the case of a nationwide rollout, the nonfunctional requirements' scaleability, adaptability as well as real-time capability are important for ITS solutions \cite{Hao.2020,Ogie.2017}. These high demands are big challenges in the software architecture: centralized software solutions (e.g. with a cloud system) are available for ITS, but they have numerous disadvantages in scalability, energy efficiency, and failure tolerance. A scalable architecture to solve the mentioned problems could be a distributed approach. According to \cite{Petitti.2016}, this approach for intelligent sensor networks was not available, or was not adequately researched. Indeed, with the use of edge computing, the papers \cite{Datta.,AmilcareFrancescoSantamaria.2018} presented distributed approaches for ITS architectures. Furthermore, the mentioned ICSI test bed has also focused on highly scalable distributed ITS \cite{Osaba.2016}. However, this test bed only analyzed specific situations. It has not processed digital twins in real-time. To the best of our knowledge, it is still unclear how this distributed system performs in a high data density environment with digital twins. For this reason, further research in scalability, as well as adaptability in real-world test beds is warranted.

%% file: diagram_sensor_trend_cumulative.tex
\begin{tikzpicture}
	\begin{axis}[
		title = {The Diversity of Sensors},
		ylabel= {Cumulative Number of ITS},
		xmin = 1986, ymin = 0,
		xtick = {1986, 1988, ..., 2021},
		ytick = {0, 10, ..., 60},
		scaled ticks=false,
		ticklabel style={
			/pgf/number format/fixed,
			/pgf/number format/1000 sep={}
		},
		x tick label style={rotate=60},
		legend style={at={(0.03,0.5)},anchor=west},
		scale only axis,
		width=0.92\textwidth,
		height=5cm,
		]
		\addplot coordinates {
			(1986, 0)
			(1987, 0)
			(1988, 0)
			(1989, 0)
			(1990, 0)
			(1991, 0)
			(1992, 0)
			(1993, 0)
			(1994, 0)
			(1995, 0)
			(1996, 0)
			(1997, 0)
			(1998, 0)
			(1999, 0)
			(2000, 0)
			(2001, 0)
			(2002, 0)
			(2003, 0)
			(2004, 0)
			(2005, 0)
			(2006, 1)
			(2007, 1)
			(2008, 1)
			(2009, 2)
			(2010, 2)
			(2011, 2)
			(2012, 3)
			(2013, 4)
			(2014, 6)
			(2015, 8)
			(2016, 14)
			(2017, 19)
			(2018, 22)
			(2019, 23)
			(2020, 24)
			(2021, 25)
		};
	\addlegendentry{Camera}
	
	\addplot coordinates {
			(1986, 0)
			(1987, 0)
			(1988, 0)
			(1989, 0)
			(1990, 0)
			(1991, 0)
			(1992, 0)
			(1993, 0)
			(1994, 0)
			(1995, 0)
			(1996, 0)
			(1997, 0)
			(1998, 0)
			(1999, 0)
			(2000, 0)
			(2001, 0)
			(2002, 0)
			(2003, 0)
			(2004, 0)
			(2005, 0)
			(2006, 0)
			(2007, 0)
			(2008, 0)
			(2009, 0)
			(2010, 0)
			(2011, 0)
			(2012, 0)
			(2013, 0)
			(2014, 1)
			(2015, 1)
			(2016, 2)
			(2017, 4)
			(2018, 6)
			(2019, 7)
			(2020, 7)
			(2021, 8)
	};
	\addlegendentry{Radar}
	
	\addplot coordinates {
			(1986, 0)
			(1987, 0)
			(1988, 0)
			(1989, 0)
			(1990, 0)
			(1991, 0)
			(1992, 0)
			(1993, 0)
			(1994, 0)
			(1995, 0)
			(1996, 0)
			(1997, 0)
			(1998, 0)
			(1999, 0)
			(2000, 0)
			(2001, 0)
			(2002, 0)
			(2003, 0)
			(2004, 0)
			(2005, 0)
			(2006, 1)
			(2007, 1)
			(2008, 1)
			(2009, 2)
			(2010, 2)
			(2011, 2)
			(2012, 2)
			(2013, 2)
			(2014, 3)
			(2015, 4)
			(2016, 5)
			(2017, 8)
			(2018, 10)
			(2019, 11)
			(2020, 11)
			(2021, 12)
	};
	\addlegendentry{Lidar}
	
	\addplot coordinates {
			(1986, 0)
			(1987, 0)
			(1988, 0)
			(1989, 0)
			(1990, 0)
			(1991, 0)
			(1992, 0)
			(1993, 0)
			(1994, 0)
			(1995, 0)
			(1996, 0)
			(1997, 0)
			(1998, 1)
			(1999, 1)
			(2000, 1)
			(2001, 1)
			(2002, 1)
			(2003, 1)
			(2004, 1)
			(2005, 1)
			(2006, 1)
			(2007, 1)
			(2008, 1)
			(2009, 1)
			(2010, 1)
			(2011, 1)
			(2012, 1)
			(2013, 1)
			(2014, 2)
			(2015, 2)
			(2016, 3)
			(2017, 4)
			(2018, 4)
			(2019, 4)
			(2020, 4)
			(2021, 4)
	};
	\addlegendentry{Induction Loop}
	
	\addplot coordinates {
			(1986, 0)
			(1987, 0)
			(1988, 0)
			(1989, 0)
			(1990, 0)
			(1991, 0)
			(1992, 0)
			(1993, 0)
			(1994, 0)
			(1995, 0)
			(1996, 0)
			(1997, 0)
			(1998, 0)
			(1999, 0)
			(2000, 0)
			(2001, 0)
			(2002, 0)
			(2003, 0)
			(2004, 0)
			(2005, 0)
			(2006, 1)
			(2007, 1)
			(2008, 1)
			(2009, 1)
			(2010, 1)
			(2011, 1)
			(2012, 2)
			(2013, 2)
			(2014, 3)
			(2015, 4)
			(2016, 8)
			(2017, 10)
			(2018, 12)
			(2019, 12)
			(2020, 12)
			(2021, 13)
	};
	\addlegendentry{Special Inputs}
	
	\addplot coordinates {
		(1986, 2)
			(1987, 2)
			(1988, 2)
			(1989, 2)
			(1990, 2)
			(1991, 2)
			(1992, 2)
			(1993, 2)
			(1994, 2)
			(1995, 2)
			(1996, 2)
			(1997, 2)
			(1998, 2)
			(1999, 2)
			(2000, 2)
			(2001, 2)
			(2002, 2)
			(2003, 2)
			(2004, 2)
			(2005, 2)
			(2006, 3)
			(2007, 3)
			(2008, 3)
			(2009, 4)
			(2010, 4)
			(2011, 4)
			(2012, 6)
			(2013, 7)
			(2014, 12)
			(2015, 14)
			(2016, 25)
			(2017, 34)
			(2018, 43)
			(2019, 47)
			(2020, 48)
			(2021, 50)
	};
	\addlegendentry{V2X}
		
	\end{axis}
\end{tikzpicture}

%% file: diagram_value_added_service_trend_cumulative.tex
\begin{tikzpicture}
	\begin{axis}[
		title = {The Cumulative Trend of Value-Added Services},
		ylabel= {Cumulative Number of ITS},
		xmin = 1986, ymin = 0,
		xtick = {1986, 1988, ..., 2021},
		ytick = {0, 2, ..., 60},
		scaled ticks=false,
		ticklabel style={
			/pgf/number format/fixed,
			/pgf/number format/1000 sep={}
		},
		x tick label style={rotate=60},
		legend style={at={(0.03,0.5)},anchor=west},
		scale only axis,
		width=0.92\textwidth,
		height=5cm,
		]

		\addplot coordinates {
			(1986, 0)
			(1987, 0)
			(1988, 0)
			(1989, 0)
			(1990, 0)
			(1991, 0)
			(1992, 0)
			(1993, 0)
			(1994, 0)
			(1995, 0)
			(1996, 0)
			(1997, 0)
			(1998, 0)
			(1999, 0)
			(2000, 0)
			(2001, 0)
			(2002, 0)
			(2003, 0)
			(2004, 0)
			(2005, 0)
			(2006, 1)
			(2007, 1)
			(2008, 1)
			(2009, 1)
			(2010, 1)
			(2011, 1)
			(2012, 1)
			(2013, 1)
			(2014, 3)
			(2015, 4)
			(2016, 7)
			(2017, 9)
			(2018, 12)
			(2019, 12)
			(2020, 12)
			(2021, 13)
		};
		\addlegendentry{Warning in operational maneuvers}
		
		\addplot coordinates {
			(1986, 1)
			(1987, 1)
			(1988, 1)
			(1989, 1)
			(1990, 1)
			(1991, 1)
			(1992, 1)
			(1993, 1)
			(1994, 1)
			(1995, 1)
			(1996, 1)
			(1997, 1)
			(1998, 1)
			(1999, 1)
			(2000, 1)
			(2001, 1)
			(2002, 1)
			(2003, 1)
			(2004, 1)
			(2005, 1)
			(2006, 1)
			(2007, 1)
			(2008, 1)
			(2009, 1)
			(2010, 1)
			(2011, 1)
			(2012, 1)
			(2013, 2)
			(2014, 3)
			(2015, 3)
			(2016, 8)
			(2017, 12)
			(2018, 14)
			(2019, 15)
			(2020, 15)
			(2021, 15)
		};
		\addlegendentry{Warning in road conditions}
		
		\addplot coordinates {
			(1986, 1)
			(1987, 1)
			(1988, 1)
			(1989, 1)
			(1990, 1)
			(1991, 1)
			(1992, 1)
			(1993, 1)
			(1994, 1)
			(1995, 1)
			(1996, 1)
			(1997, 1)
			(1998, 1)
			(1999, 1)
			(2000, 1)
			(2001, 1)
			(2002, 1)
			(2003, 1)
			(2004, 1)
			(2005, 1)
			(2006, 2)
			(2007, 2)
			(2008, 2)
			(2009, 2)
			(2010, 2)
			(2011, 2)
			(2012, 2)
			(2013, 3)
			(2014, 4)
			(2015, 4)
			(2016, 7)
			(2017, 13)
			(2018, 14)
			(2019, 15)
			(2020, 15)
			(2021, 16)
		};
		\addlegendentry{Warning in traffic situations}
		
		\addplot coordinates {
			(1986, 1)
			(1987, 1)
			(1988, 1)
			(1989, 1)
			(1990, 1)
			(1991, 1)
			(1992, 1)
			(1993, 1)
			(1994, 1)
			(1995, 1)
			(1996, 1)
			(1997, 1)
			(1998, 1)
			(1999, 1)
			(2000, 1)
			(2001, 1)
			(2002, 1)
			(2003, 1)
			(2004, 1)
			(2005, 1)
			(2006, 1)
			(2007, 1)
			(2008, 1)
			(2009, 1)
			(2010, 1)
			(2011, 1)
			(2012, 1)
			(2013, 1)
			(2014, 3)
			(2015, 4)
			(2016, 7)
			(2017, 9)
			(2018, 12)
			(2019, 13)
			(2020, 13)
			(2021, 14)
		};
		\addlegendentry{Controlling of vehicles}
		
		\addplot coordinates {
			(1986, 0)
			(1987, 0)
			(1988, 0)
			(1989, 0)
			(1990, 0)
			(1991, 0)
			(1992, 0)
			(1993, 0)
			(1994, 0)
			(1995, 0)
			(1996, 0)
			(1997, 0)
			(1998, 1)
			(1999, 1)
			(2000, 1)
			(2001, 1)
			(2002, 1)
			(2003, 1)
			(2004, 1)
			(2005, 1)
			(2006, 1)
			(2007, 1)
			(2008, 1)
			(2009, 1)
			(2010, 1)
			(2011, 1)
			(2012, 2)
			(2013, 3)
			(2014, 3)
			(2015, 3)
			(2016, 8)
			(2017, 12)
			(2018, 14)
			(2019, 15)
			(2020, 15)
			(2021, 16)
		};
		\addlegendentry{Controlling of the traffic flow}
		
	\end{axis}
\end{tikzpicture}

%% file: diagram_value_added_service_trend_absolute.tex
\begin{tikzpicture}
	\begin{axis}[
		title = {The Absolute Trend of Value-Added Services},
		ylabel= {Absolute Number of ITS per Year},
		xmin = 1986, ymin = 0,
		xtick = {1986, 1988, ..., 2021},
		ytick = {0, 2, ..., 60},
		scaled ticks=false,
		ticklabel style={
			/pgf/number format/fixed,
			/pgf/number format/1000 sep={}
		},
		x tick label style={rotate=60},
		legend style={at={(0.03,0.5)},anchor=west},
		scale only axis,
		width=0.92\textwidth,
		height=5cm,
		]

		\addplot coordinates {
			(1986, 0)
			(1987, 0)
			(1988, 0)
			(1989, 0)
			(1990, 0)
			(1991, 0)
			(1992, 0)
			(1993, 0)
			(1994, 0)
			(1995, 0)
			(1996, 0)
			(1997, 0)
			(1998, 0)
			(1999, 0)
			(2000, 0)
			(2001, 0)
			(2002, 0)
			(2003, 0)
			(2004, 0)
			(2005, 0)
			(2006, 1)
			(2007, 0)
			(2008, 0)
			(2009, 0)
			(2010, 0)
			(2011, 0)
			(2012, 0)
			(2013, 0)
			(2014, 2)
			(2015, 1)
			(2016, 3)
			(2017, 2)
			(2018, 3)
			(2019, 0)
			(2020, 0)
			(2021, 1)
		};
		\addlegendentry{Warning in operational maneuvers}
		
		\addplot coordinates {
			(1986, 1)
			(1987, 0)
			(1988, 0)
			(1989, 0)
			(1990, 0)
			(1991, 0)
			(1992, 0)
			(1993, 0)
			(1994, 0)
			(1995, 0)
			(1996, 0)
			(1997, 0)
			(1998, 0)
			(1999, 0)
			(2000, 0)
			(2001, 0)
			(2002, 0)
			(2003, 0)
			(2004, 0)
			(2005, 0)
			(2006, 0)
			(2007, 0)
			(2008, 0)
			(2009, 0)
			(2010, 0)
			(2011, 0)
			(2012, 0)
			(2013, 1)
			(2014, 1)
			(2015, 0)
			(2016, 5)
			(2017, 4)
			(2018, 2)
			(2019, 1)
			(2020, 0)
			(2021, 0)
		};
		\addlegendentry{Warning in road conditions}
		
		\addplot coordinates {
			(1986, 1)
			(1987, 0)
			(1988, 0)
			(1989, 0)
			(1990, 0)
			(1991, 0)
			(1992, 0)
			(1993, 0)
			(1994, 0)
			(1995, 0)
			(1996, 0)
			(1997, 0)
			(1998, 0)
			(1999, 0)
			(2000, 0)
			(2001, 0)
			(2002, 0)
			(2003, 0)
			(2004, 0)
			(2005, 0)
			(2006, 1)
			(2007, 0)
			(2008, 0)
			(2009, 0)
			(2010, 0)
			(2011, 0)
			(2012, 0)
			(2013, 1)
			(2014, 1)
			(2015, 0)
			(2016, 3)
			(2017, 6)
			(2018, 1)
			(2019, 1)
			(2020, 0)
			(2021, 1)
		};
		\addlegendentry{Warning in traffic situations}
		
		\addplot coordinates {
			(1986, 1)
			(1987, 0)
			(1988, 0)
			(1989, 0)
			(1990, 0)
			(1991, 0)
			(1992, 0)
			(1993, 0)
			(1994, 0)
			(1995, 0)
			(1996, 0)
			(1997, 0)
			(1998, 0)
			(1999, 0)
			(2000, 0)
			(2001, 0)
			(2002, 0)
			(2003, 0)
			(2004, 0)
			(2005, 0)
			(2006, 0)
			(2007, 0)
			(2008, 0)
			(2009, 0)
			(2010, 0)
			(2011, 0)
			(2012, 0)
			(2013, 0)
			(2014, 2)
			(2015, 1)
			(2016, 3)
			(2017, 2)
			(2018, 3)
			(2019, 1)
			(2020, 0)
			(2021, 1)
		};
		\addlegendentry{Controlling of vehicles}
		
		\addplot coordinates {
			(1986, 0)
			(1987, 0)
			(1988, 0)
			(1989, 0)
			(1990, 0)
			(1991, 0)
			(1992, 0)
			(1993, 0)
			(1994, 0)
			(1995, 0)
			(1996, 0)
			(1997, 0)
			(1998, 1)
			(1999, 0)
			(2000, 0)
			(2001, 0)
			(2002, 0)
			(2003, 0)
			(2004, 0)
			(2005, 0)
			(2006, 0)
			(2007, 0)
			(2008, 0)
			(2009, 0)
			(2010, 0)
			(2011, 0)
			(2012, 1)
			(2013, 1)
			(2014, 0)
			(2015, 0)
			(2016, 5)
			(2017, 4)
			(2018, 2)
			(2019, 1)
			(2020, 0)
			(2021, 1)
		};
		\addlegendentry{Controlling of the traffic flow}
		
	\end{axis}
\end{tikzpicture}

%% file: conclusion.tex
\section{Conclusion}
\label{section:conclusion}
Within the scope of this work, 357 papers and documents were collected in a systematic review to provide a summary of current intelligent transportation systems. After passing the filtering, 127 documents with 52 test bed projects were analyzed. In this survey, we have highlighted the fascinating development in the field of ITS using external infrastructure from the `Early Days' until the `Creation of Digital Twins'.

We have highlighted the immense increase in the complexity and functionality of ITS as well as the outstanding challenges. The current systems can deliver highly accurate information about individuals in traffic situations in real-time. In the field of ITS, this development is a great success. However, improvements in the reliability, scalability, and security of the overall system are still needed. For this reason, further research in the ITS should focus on more reliable perception of the traffic using modern sensors, plug-and-play mechanisms as well as secure real-time distribution in a decentralized manner of a large amount of data. Furthermore, to ensure such systems are viable, services with a real added value for the general public with autonomous as well as non-autonomous cars have to be developed. 

In summary, it can be seen that the development of intelligent transportation systems using external infrastructure in proceeding in the correct direction of autonomous driving. However, a few questions remain outstanding before a comprehensive roll-out can occur.